\documentclass[a4paper,fleqn]{cas-sc}

\usepackage{natbib}

\usepackage{mathtools} 

\usepackage[T1]{fontenc}
\usepackage{graphicx}
\usepackage{hyperref}
\usepackage{amsmath,amsthm,amssymb}
\usepackage{amsfonts}

\usepackage{algorithm}%
\usepackage{algorithmicx}%
\usepackage{algpseudocode}%
\usepackage{subfig}
\usepackage[labelfont=bf]{caption}
\usepackage{xcolor}
\usepackage{array}
\usepackage{multirow}
\usepackage{enumitem}
\usepackage{rotating}
\usepackage{color}
\usepackage{dsfont}
\usepackage{booktabs} 

\usepackage{placeins} 
\usepackage{subfig}
\usepackage{textgreek}
\usepackage{lineno}

\newcommand{\revisionchange}[1]{#1}
\newenvironment{revisionsection}{}{}

\DeclareMathOperator*{\argmax}{arg\,max}

\begin{document}

\shorttitle{SoilNet}    

\shortauthors{Singh \& Chiaburu et. al.} 

\title[mode=title]{SoilNet: A Multimodal Multitask Model for Hierarchical Classification of Soil Horizons}

\author[1]{Vipin Singh}[orcid=0000-0002-4472-8285]
\cormark[1] 
\fnmark[1] 
\ead{vipin.singh@bht-berlin.de} 
\credit{Writing - original draft, Conceptualization, Methodology, Investigation, Formal analysis, Software, Visualization}

\author[1]{Teodor Chiaburu}[orcid=0009-0009-5336-2455]
\fnmark[1] 
\ead{chiaburu.teodor@bht-berlin.de} 
\credit{Writing - original draft, Conceptualization, Methodology, Investigation, Formal analysis, Software, Visualization}

\author[2]{Einar Eberhardt}[orcid=0000-0003-1499-618X]
\ead{einar.eberhardt@bgr.de} 
\credit{Writing – review \& editing, Data curation, Validation}

\author[2]{Stefan Broda}[orcid=0000-0001-6858-6368]
\ead{stefan.broda@bgr.de} 
\credit{Writing – review \& editing, Validation, Funding acquisition}

\author[1]{\revisionchange{Joey Prüssing}}
\ead{s85955@bht-berlin.de}
\credit{\revisionchange{Software, Visualization}}

\author[1]{Frank Haußer}[orcid=0000-0002-8060-8897]
\fnmark[2] 
\ead{frank.hausser@bht-berlin.de} 
\credit{Writing – review \& editing, Conceptualization, Validation, Supervision}

\author[1,3]{Felix Bießmann}[orcid=0000-0002-3422-1026]
\fnmark[2] 
\ead{felix.biessmann@bht-berlin.de} 
\credit{Writing – review \& editing, Conceptualization, Funding acquisition, Validation, Supervision}

\affiliation[1]{organization={Berliner Hochschule für Technik, Luxemburger Straße 10, 13353 Berlin, Germany}}
\affiliation[2]{organization={Bundesanstalt für Geowissenschaften und Rohstoffe, Stilleweg 2, 30655 Hannover, Germany}}
\affiliation[3]{organization={Einstein Center Digital Future, Wilhelmstrasse 67, 10117 Berlin, Germany}}
\cortext[1]{Corresponding author.}
\fntext[1]{Equal contribution.}
\fntext[2]{Equal supervision.}

\begin{abstract}
Recent advances in artificial intelligence (AI), in particular foundation models, have improved the state of the art in many application domains including geosciences. Some specific problems, however, could not benefit from this progress yet. Soil horizon classification, for instance, remains challenging because of its multimodal and multitask characteristics and a complex hierarchically structured label taxonomy. Accurate classification of soil horizons is crucial for monitoring soil condition, which directly impacts agricultural productivity, food security, ecosystem stability and climate resilience. 
In this work, we propose \textit{SoilNet} - a multimodal multitask model to tackle this problem through a structured modularized pipeline. In contrast to omnipurpose AI foundation models, our approach is designed to be inherently transparent by following the task structure human experts developed for solving this challenging annotation task. The proposed approach integrates image data and geotemporal metadata to first predict depth markers, segmenting the soil profile into horizon candidates. Each segment is characterized by a set of horizon-specific morphological features. Finally, horizon labels are predicted based on the multimodal concatenated feature vector, leveraging a graph-based label representation to account for the complex hierarchical relationships among soil horizons. Our method is designed to address complex hierarchical classification, where the number of possible labels is very large, imbalanced and non-trivially structured. We demonstrate the effectiveness of our approach on a real-world soil profile dataset \revisionchange{and a comprehensive user study with domain experts. Our empirical evaluations demonstrate that SoilNet reliably predicts soil horizons that are plausible and accurate. User study results indicate that SoilNet achieves predictive performance on par with or better than that of human experts in soil horizon classification.} All code and experiments can be found in our repository:
\href{https://github.com/calgo-lab/BGR/}{https://github.com/calgo-lab/BGR/}.

\end{abstract}



\maketitle              
\section{Introduction}\label{sec_intro}

Soils are essential to life on Earth, serving as the foundation for ecosystems, agriculture and water filtration. They support plant growth, regulate the carbon cycle and influence biodiversity, making soil condition a direct determinant of food security, environmental resilience and public well-being~\citep{desertification}. Assessing soil condition over larger areas is, therefore, critical. 
However, obtaining soil data is laborious and more or less point-related. While huge efforts have been made to develop robust regionalization methodologies in digital soil mapping~\citep{lagacherieDigitalSoilMappingSOTA, mcbratneyDigitalSoilMapping, wadouxMachineLearningDigitalSoilMapping}, there are fewer ideas about how to improve the quantity and quality of soil profile data that form the basis for the application of such methodologies with justifiable costs and in due time. Descriptive soil data are often used for decades, yet they are intrinsically linked to the underlying – and often outdated - description guidelines, the applied soil classification, and, not least, the expertise of the profile author. The challenges of updating the nomenclature and quality assessment of existing soil profile data as well as improving the data basis, e.g. by drawing on citizen science approaches, highlight the need to engage objective elements in data acquisition beside the classic profile description.



\begin{figure*}[pos=h]  
    \centering
    \includegraphics[width=0.8\linewidth]{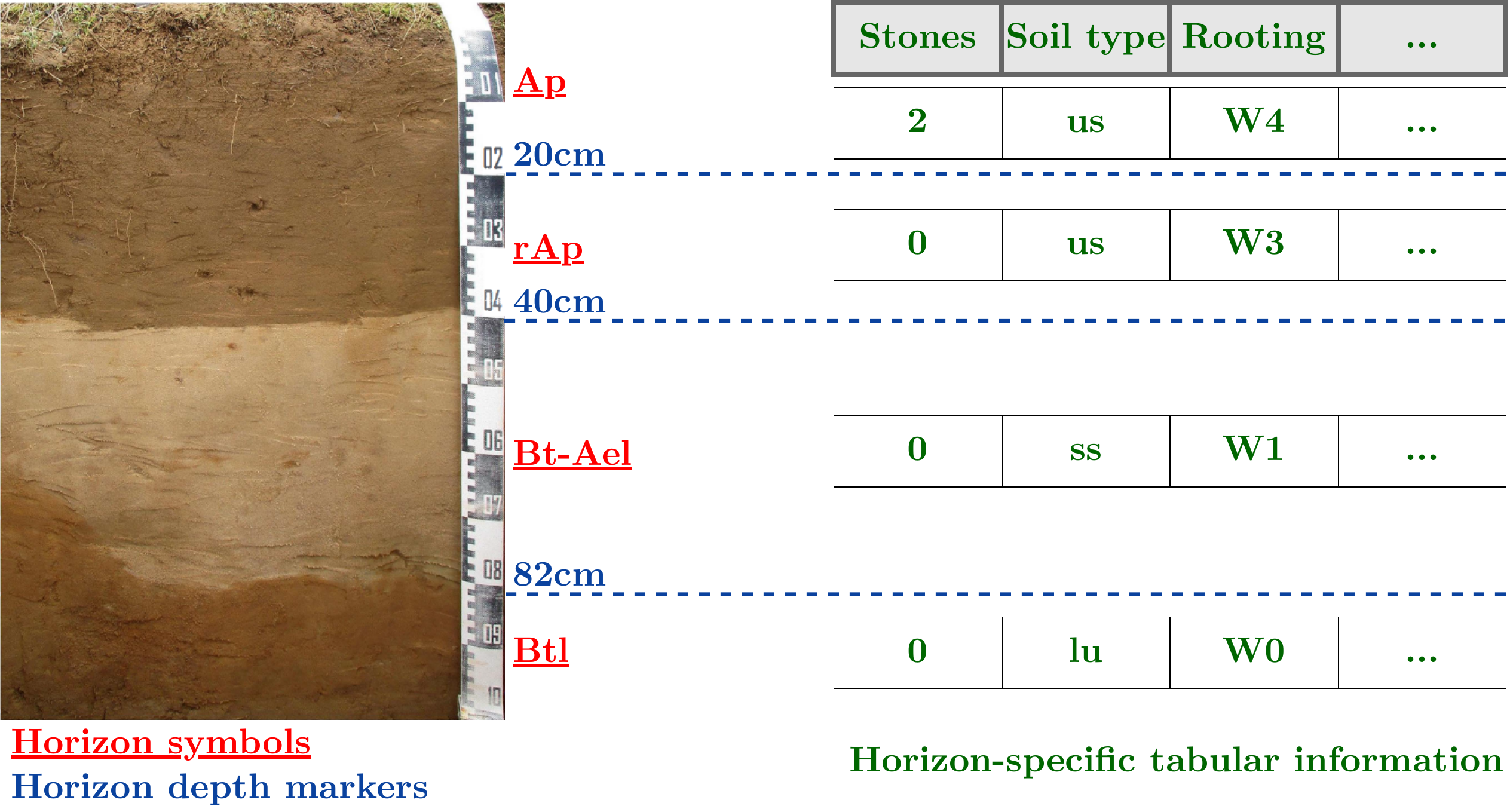}
    \caption{\textbf{Example of an annotated soil profile.} The ruler on the right margin marks the 1-meter depth of the photographic probe. 
    The soil profile has 4 horizons, with annotated labels (in red), segment borders (in blue) and morphological properties (in green). Soil images and data provided by: \cite{soil_fotos, soil_bgr}.
    }
    \label{fig:profile_example}
\end{figure*}

A key aspect of soil condition assessment is the characterization of the soil's vertical and lateral structure, i.e. the \textit{soil horizons}~\citep{HARTEMINK2020125, iussWorldReferenceSoilResources}. These are classified based on their physical, chemical and biological properties, providing essential insights into soil properties, formation processes and functions. Accurately identifying and classifying soil horizons is essential for soil mapping and regionalization of soil information. However, this task is inherently complex, given the highly variable nature of soil profiles across different regions and the intricate dependencies between horizon types. As illustrated in \autoref{fig:profile_example} and detailed in \autoref{sec:llms}, such expert annotations are highly specialized and up until now remain difficult to automate effectively  using generic prompting approaches with large language models (LLMs).

Across Europe, soil monitoring procedures, including sampling strategies or soil properties measured, differ from country to country; an extensive comparable study is provided in \cite{lucas}. Soil horizon classification, particularly the German categorization~\citep{kartieranleitung5} from the dataset used here, follows roughly a hierarchical taxonomy, but a more detailed view reveals that soil horizons exhibit more complex characteristics such as multiple parent-child relationships. To preserve these complex relationships, the soil horizon taxonomy, hence, should be represented as a graph structure rather than a simple tree. A single horizon can share properties with multiple others and transitional layers further complicate the classification process. This complexity, combined with the class imbalance in real-world soil datasets, makes automated horizon classification particularly challenging.

To address these challenges, we introduce \textit{SoilNet}, a novel multimodal framework for automated soil horizon denomination. Our approach sequentially predicts depth markers based on image data and geotemporal metadata e.g. geographical location, month, relief type, effectively segmenting soil profiles into meaningful horizon intervals. Each segmented region is then described by a set of physical/morphological soil features e.g. Munsell color reading, humus content or rooting pattern.
A key aspect of our method is the use of graph-based embeddings to represent the  horizon labels, capturing their complex interdependencies beyond a rigid hierarchical taxonomy. 


The main contributions of our work are as follows:
\begin{enumerate}
    \item We formulate the soil horizon classification task as a structured, multimodal, multitask problem, decomposed into a three-stage pipeline that mirrors the expert decision-making process in pedology:
    \begin{enumerate}
        \item \textbf{Task 1 – Segmentation}: We predict a variable-length sequence of \textit{depth markers} that segment the soil profile image into sequential horizon candidates.
        \item \textbf{Task 2 – Morphological Feature Prediction}: For each predicted segment, we estimate a set of \textit{tabular morphological properties} (e.g., Munsell soil color, humus content), serving as a reproduction of the horizons' metadata.
        \item \textbf{Task 3 – Horizon Classification}: Finally, we classify each horizon segment using \textit{hierarchical label embeddings}, accounting for the complex taxonomy of soil horizons.
    \end{enumerate}
    \item We introduce \textbf{SoilNet}, a novel end-to-end training and inference framework that integrates image and tabular modalities, sequential segmentation, intermediate tabular predictors and graph-based label reasoning. The model is modular by design, enabling transparent evaluation of each stage and facilitating adaptation to similar hierarchical segmentation and classification tasks in other domains.
\end{enumerate}


\section{Related Work}\label{sec_relwork}
\revisionchange{To contextualize our methodological choices, we review prior approaches and put our work within the broader landscape of Machine Learning applications in soil science. We group the literature into four categories: Automatic Soil Classification, Sequential Segmentation, Hierarchical Classification and Multimodal/Multitask Models. We highlight common limitations, including simplified label structures and reliance on input from a single modality, which our approach aims to overcome.}

\paragraph{Automatic Soil Classification.} Developing automated Machine Learning (ML) methods for classifying soils or predicting various soil properties has been an active field of research throughout the past years. The study in~\cite{acid_sulfate} evaluates Convolutional Neural Networks (CNNs) for a binary classification of acid sulfate soils. \revisionchange{In their results the CNNs outperform their Random Forest baseline, reporting an accuracy of 68\%.}
The research in~\cite{soil_sprectroscopy, soil_sprectroscopy_2} combines soil spectral and/or hyperspectral information with standard ML methods like SVMs or Random Forests\revisionchange{, applied to soil horizons and suborders that are spectrally well separated, a fact acknowledged by the authors and reflected in their high accuracies up to 97\%}.~\cite{soil_cnn} train CNNs to classify soil types based on their ground surface color\revisionchange{, achieving 97\% accuracy on a simple two-class dataset (red soils vs. black soils)}. ~\cite{soils_lombardia} analyze the spatial distribution of soil organic carbon using ML with Residual Kriging~\citep{resid_kriging}\revisionchange{, where Residual Kriging improves prediction performance compared to ML models alone}.~\cite{soil_selfsupervised_vit} train a Vision Transformer in a self-supervised fashion to predict various soil characteristics in drylands based on satellite image data. An in-depth review of the current ML strategies used for soil classification can be found in~\cite{soil_review1,soil_review2}. While valid approaches, we note that all methods presented in these reviews work solely with image data and the label space is much simpler (no more than 10 classes) than the one considered for our data.

\paragraph{Sequential Segmentation.} Modeling sequential outputs, where the prediction at each step depends on previous outputs and contextual cues is highly relevant to our task of sequentially segmenting soil profile images from top to bottom into horizon boundaries. A widely adopted solution in such settings is the use of Long Short-Term Memory (LSTM) networks~\citep{lstm}. More recently, \textit{Cross Attention} mechanisms within Transformer architectures have emerged as a powerful alternative for sequence modeling~\citep{crossvit,stack_transformer,bixt,cat}. In a similar ecological use-case,~\cite{root_segmentation} apply various Deep Learning techniques to segment fine roots from time series images of soil profiles, in order to investigate the dynamics of fine roots areas over time, while~\cite{soil_survey} apply semantic segmentation \revisionchange{via a nested U-Net} to directly localize and classify soil horizons into main symbols A, B, C\revisionchange{, achieving a mean intersection over union (IoU) of 77\% and a pixel accuracy of 83\%.} \revisionchange{An earlier attempt at automated horizon segmentation is presented in~\cite{soil_horizon_kmeans}, which applies k-means clustering to color and texture features to segment master horizons, though relying solely on low-level image cues.} In a different application scenario (medicine), but with comparable sequential structure, the authors in~\cite{seg_retina} segment retinal layers in tomographic images with CNN-based models by predicting the contours of their boundaries.

\paragraph{Hierarchical Classification.} Hierarchical structures are a common characteristic across a wide range of ML tasks. In computer vision, efforts to incorporate class hierarchies into image classifiers typically fall into three categories~\citep{cls_hierarchies}. The first includes \textit{label/graph embedding} approaches~\citep{embeddings_jena,nickelPoincareEmbeddingsLearning2017}, which represent labels as vectors in a continuous space to capture semantic relationships (typically derived from a tree or graph). The second group focuses on \textit{hierarchy-aware loss functions}~\citep{cls_hierarchies,10k_categories}, which encourage predictions that are consistent with the underlying class taxonomy. The third category involves \textit{hierarchical architectures}~\citep{bcnn,hdcnn} that integrate the hierarchical structure directly into the network design. Recent advancements explore hierarchical segmentation~\citep{liDeepHierarchicalSemantic2022}, combining the aforementioned strategies to directly enforce the label hierarchy on the segmentation task.

\paragraph{Multimodal/Multitask Models.} Recent work in natural language processing has led to the development of large language models (LLMs) that serve as general-purpose multitask solvers, capable of addressing a wide range of linguistic tasks in both zero-shot and few-shot scenarios~\citep{llm_fewshot}. Efforts to extend these models into multimodal domains have followed three main directions. One line of work focuses on unifying vision and language inputs within a single architecture to enable perception and reasoning across different modalities~\citep{flamingo,kosmos}. Another direction explores the integration of external tools and models into LLM pipelines, allowing them to dynamically invoke specialized modules such as visual encoders or programmatic solvers~\citep{vipergpt,visprog}. A third direction treats the LLM as a centralized controller responsible for orchestrating a collection of expert models across modalities~\citep{hugginggpt,olympus}. Rather than solving tasks directly, the LLM interprets user intent, selects appropriate external models based on their capabilities and manages inter-model communication.

We note that, to the best of our knowledge, none of the current ML approaches leverages multimodal input for classifying soil horizons (the typical input is images). In contrast, our work combines visual input with tabular \revisionchange{descriptive horizon data}.
This offers a much richer signal to the classifier. A more in-depth discussion is provided in \autoref{subsec:soilnet}.

\section{Methods}\label{sec_meth}

In this section we describe our data along with preprocessing steps, the overall workflow of our proposed approach followed by a more in-depth description of the key components in the SoilNet architecture.

\subsection{Data}\label{subsec:data}

Our image-tabular dataset was collected and curated by the Thünen Institute~\citep{soil_fotos}. This dataset is, to the best of our knowledge, 
the most comprehensive multimodal dataset in the context of soil horizon classification literature. A detailed overview of other datasets used for developing automatic soil classifiers can be found in~\cite{soil_review1}. In the following, we describe the data and the preprocessing methods. 
For the time being, the full dataset cannot be made publicly available. For a detailed description of the data acquisition and annotation procedures we refer the reader to~\cite{kartieranleitung5}.

The visual modality in our dataset consists of \textbf{3349 soil profile images} collected by soil surveyors during fieldwork. A rectangular pit approximately 1 meter deep was excavated at each sampling location to expose a vertical cross-section of the soil. Researchers then captured frontal images of the exposed soil wall, where natural stratification into horizons is visible through color, texture and structural variation (see example in \autoref{fig:profile_example}). These photographs provide a consistent top-down view of the soil profile, with the surface typically aligned at the top of the frame and the bottom corresponding to a depth of around -1 meter. 
From the original profile images we have removed the ruler with SAM~\citep{sam} and, whenever necessary, the sky background visible above the ground with a standard thresholding algorithm. This way, all the resulting images correspond from top to bottom to the 1-meter-range of the \revisionchange{soil profile}.

The profile images are accompanied by a set of additional annotations, stored in \textbf{tabular form}. These include some that are horizon-specific and others that describe the whole \revisionchange{soil profile}, such as associated geotemporal data. In total, the dataset contains data for \textbf{13621 horizons} and \textbf{1218 horizon classes} (clustered as 99 classes, see \autoref{sec:label_cluster}). We emphasize that different soil profile images may exhibit a different number of horizons (every sample in the dataset has at least 2 horizons and at most 8). A detailed description of the tabular data is given below:


\paragraph{Geotemporal Information.} For each soil sample, the geographical location (as latitude/longitude coordinates) and the date (month and year) were recorded. Seasonality and location are known to influence properties such as soil moisture and hence color. Further information (derivable from time and position) was stored as well e.g. type of relief form, soil climate zone, peat thickness, groundwater level and so on. These properties describe the excavation spot as a whole and are not horizon specific.

\paragraph{Depth Markers.} The boundaries between the soil horizons are described by vertical markers stored as integers from 0 cm (ground level) to 100 cm (the deepest level of the excavated sample), see~\autoref{fig:profile_example}. Please note that the depth markers are approximations of the true boundaries between the horizons, which are often non-linear and irregular.

\paragraph{Morphological (Horizon-Specific) Information.} After segmenting the soil into horizons and upon further scrutiny and tests, the soil surveyors annotate each horizon with an additional set of morphological/physical characteristics. There are 6 such features in our dataset: 
\begin{itemize}
    \item \textit{Percentage of coarse fragments} [numerical]: To roughly estimate the gravel presence, the surveyor assessed the coarse fragment content within each soil horizon. Their numbers range from 0 to 100 in our dataset.
    
    \item \textit{Soil texture} [categorical]: The classification of soil texture is based on the relative proportions of three fine soil fractions: \textit{sand, silt} and \textit{clay}. Depending on which fraction dominates, soils are roughly categorized as sandy, silty\revisionchange{, loamy} or clayey \revisionchange{according to the German texture classification~\citep{kühn_eberhardt_soil_texture}}. For bog and fen soils, peat type data were used instead. To balance the  distribution for very rare categories, we have clustered the 58 soil texture and peat classes into their main coarse groups according to the rules provided in \cite{kartieranleitung5}, which resulted in 17 classes left in our dataset. 
    
    \item \textit{Soil color} [categorical]: Soil surveyors annotate soil horizon color
    using the standardized \textbf{Munsell color system}, with the three components: \textit{hue}, \textit{value} (lightness) and \textit{chroma} (intensity).
    To reduce the complexity of the color label space and to balance the distribution of categories in color space, we discretized the value and chroma axes into 2-by-2 grids, grouping them into coarser intervals. By doing this, we arrived at 74 color classes from 254. \revisionchange{Alternatively, one could convert the Munsell scale into CIE-L*a*b* scale, which was shown to be particularly suitable for statistical analysis~\citep{cie_lab}.}
    
    \item \textit{Carbonate content} [categorical]: Carbonate content refers to the presence of calcium carbonate ($\text{CaCO}_3$) and calcium-magnesium carbonate ($\text{CaMg(CO}_3)_2$). In our dataset, this property is discretized into levels ranging from \textit{c0} (no detectable carbonate or below 2\% of mass) to \textit{c6} (very high carbonate concentration,  between 50-75\% of mass).
    
    \item \textit{Humus content} [categorical]: Humus content reflects the amount of decomposed organic matter present in a soil horizon. Humus forms through the microbial breakdown of plant and animal residues and plays a key role in soil fertility and structure. Our dataset categorizes this property using a discrete scale from \textit{h0} (no visible humus) to \textit{h7} (very high humus content, more than 30\% of mass).
    
    \item \textit{Rooting} [categorical]: Rooting is measured as the average number of roots per $\text{dm}^2$ and is categorized on a discrete scale from \textit{W0} (no roots) to \textit{W6} (more than 50 roots per $\text{dm}^2$).
\end{itemize}

\paragraph{Horizon Labels.} The most central component of the tabular data is the soil horizon labels, which encode expert pedological interpretation of the morphological and compositional characteristics of each soil layer. These labels follow a domain-specific label grammar standardized in German pedology, where each label consists of a mandatory uppercase main symbol e.g. \textbf{A, B, C}, that denotes the principal genetic horizon, optionally preceded and/or followed by lowercase modifiers. Prefix letters, appearing before the main symbol e.g. \textbf{rA, ilC}, indicate geogenic or anthropogenic influences, such as stratification or human disturbance. Suffix letters, following the main symbol e.g. \textbf{Ah, Bv}, describe pedogenic features such as accumulation of clay, organic matter enrichment or carbonate presence. While these components can, in theory, be combined, not all combinations are valid; certain modifiers are restricted to specific main horizon types due to their pedological interpretation. For example, the prefix \textbf{a} cannot appear before \textbf{B} to form \textbf{aB}. Other countries employ horizon labeling systems that are similar in concept, but often less detailed~\citep{world_soils}.

In addition to singular horizon labels, the dataset also contains \textbf{complex labels}, which denote transition zones between two adjacent horizons that exhibit characteristics of both. These occur along visually or chemically gradual boundaries and are represented by combining two valid horizon labels with an operator, i. e. '+', '-', or '°', each reflecting a different type of transition~\citep{kartieranleitung5}. For simplification, we normalize all such mixtures using the '-' operator e.g. \textbf{Ah-Bv}, regardless of the original transition type. 
Note that in the German system used here, emphasis is on the second label part in such combinations; according to~\cite{iussWorldReferenceSoilResources}, the example horizon Ah-Bv would be a BA horizon in the international notation (\textbf{B}-dominant).

\begin{figure*}[pos=h]
    \centering
    \includegraphics[width=1\linewidth]{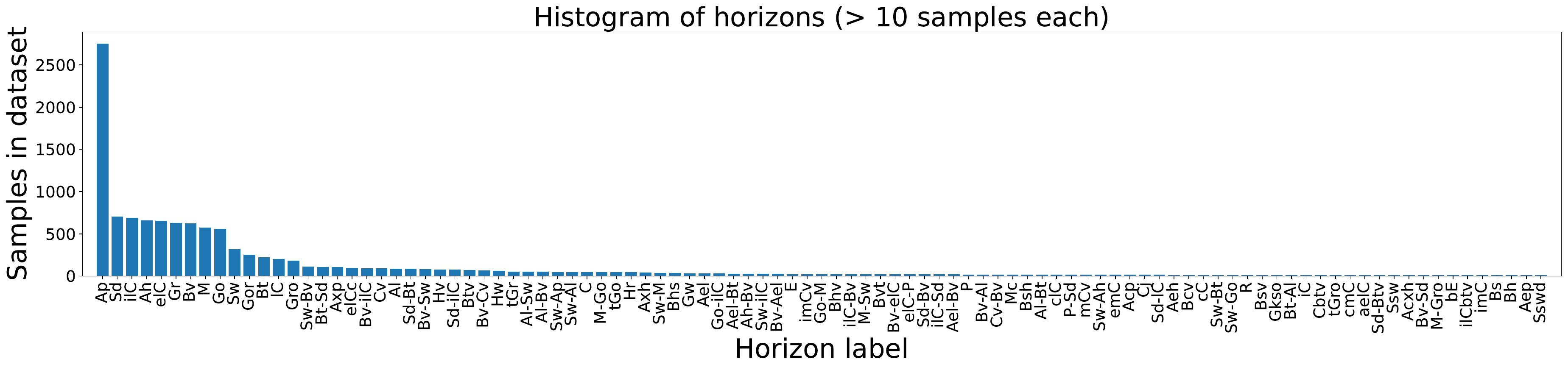}
    \caption{\textbf{Histogram of Horizons.} Distribution of horizon labels with more than 10 samples in the whole dataset (out of 13621 horizon samples in total). For training, we combine both mixture operators '-' and '+' into one single general mixture operator '-'.}
    \label{fig:horizon_histo}
\end{figure*}

With this aggregation considered, there are 1218 different horizon labels in our dataset, spread over a highly skewed distribution. These were simplified into shorter and more concise strings. From these simplified labels, we kept those that have more than 10 samples in the dataset as such; this resulted in \textbf{99 final horizon labels} - see \autoref{fig:horizon_histo} for their distribution. The remaining less frequent ones were clustered onto the most (pedologically) relevant label from the 99 ones. For details regarding this label clustering process, please consult \autoref{sec:label_cluster}.

\begin{figure*}[pos=h]
    \centering
    \includegraphics[width=1\linewidth]{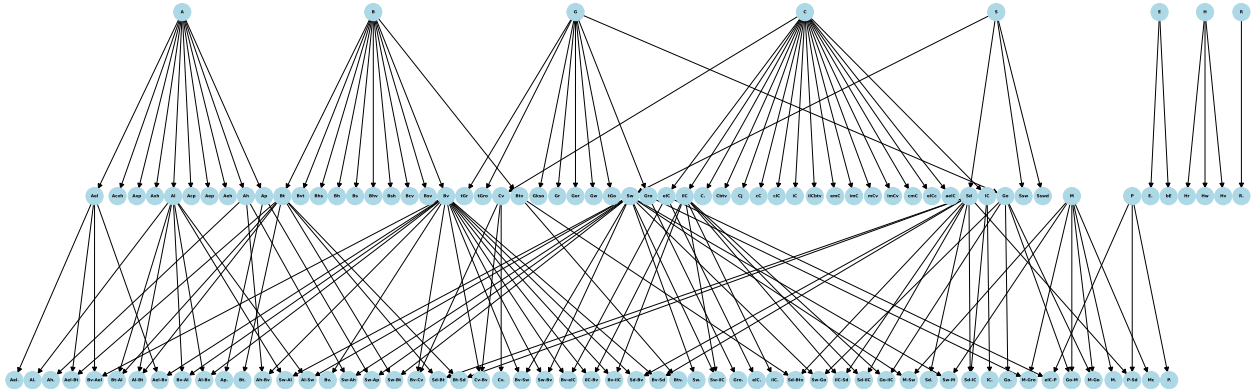}
    \caption{\textbf{Graph taxonomy of the horizon labels with more than 10 samples each.} The abstract Root node, from which the main symbols branch out, is not drawn here for better readability. All leaf nodes represent the labels present in the training set. Please note that the lowest level of the graph also contains complex nodes that derive properties from two parent nodes. Therefore, the hierarchical graph is not a tree.}
    \label{fig:graph}
\end{figure*}

The remaining 99 labels are underlain by a directed acyclic graph, visualized completely in~\autoref{fig:graph}. One particularity about the horizon label graph is that terminal nodes corresponding to the complex labels e.g. \textbf{Al-Bt} spawn from two parent nodes, one for each individual mixture member. It is this property that differentiates our taxonomic graph from a tree. 

\paragraph{Data Split.} For training our models we split the dataset into training/validation/test subsets in a proportion of 60-20-20\%. To maintain the distribution of the horizon labels and the categorical tabular horizon features used in solving Task 3 and Task 2 (see \autoref{subsec:tasks}), respectively, we applied a Multilabel Stratified Split~\citep{multistratified}. For more details regarding the preprocessing of the dataset in preparation for the training process please consult our code repository.

\subsection{Formulation of the Tasks}\label{subsec:tasks}
\begin{figure*}[pos=h]
    \centering
    \includegraphics[width=0.75\linewidth]{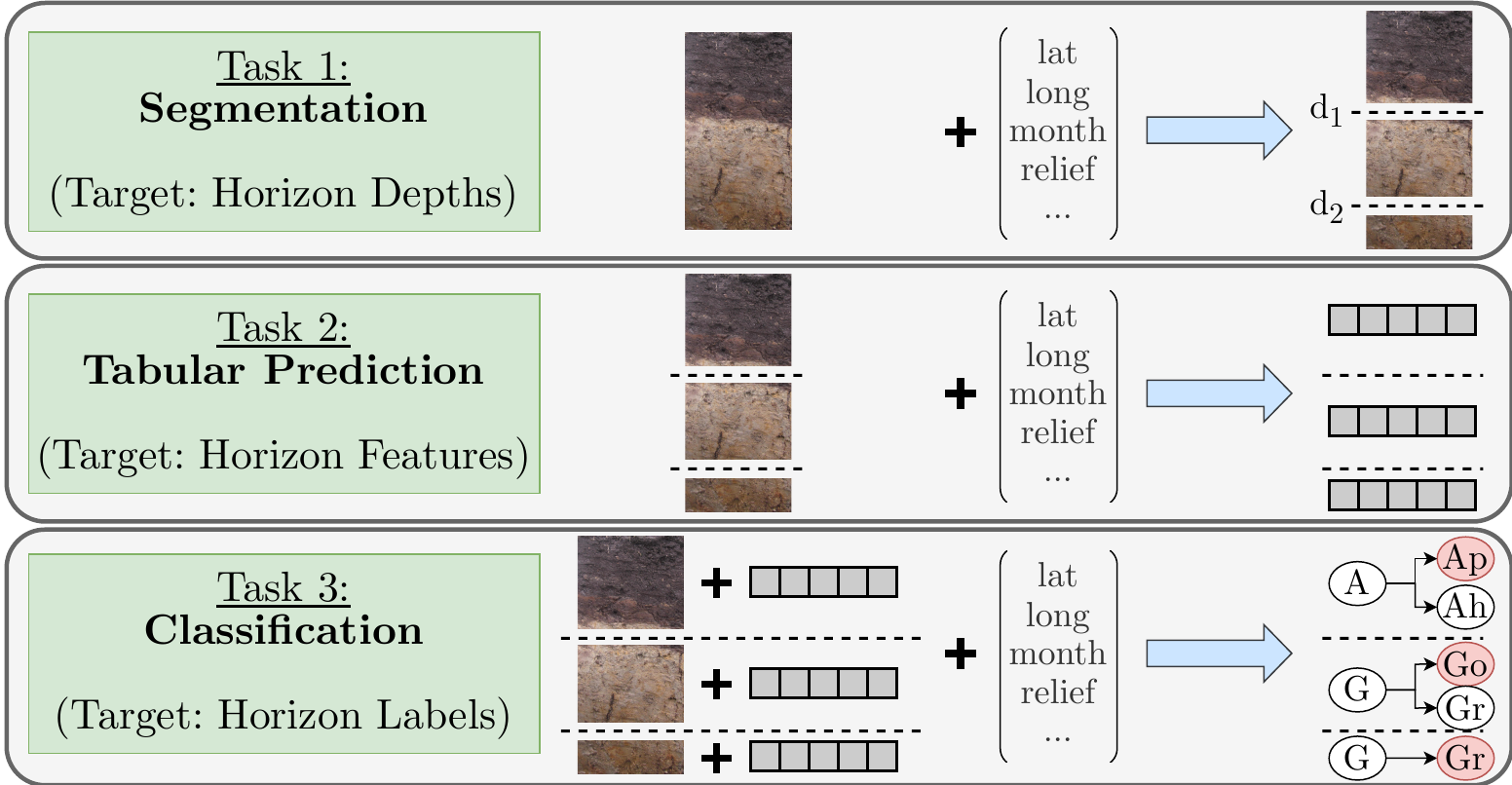}
    \caption{\textbf{Multitask Structure for Soil Horizon Classification.} \textbf{Task 1:} Soil profile images are segmented based on features extracted from the full images
    concatenated with features extracted from the geotemporal data.
    \textbf{Task 2:} Tabular morphological features are predicted based on (visual) features extracted from the segments concatenated with the geotemporal features (one set of tabular features per segment). \textbf{Task 3:} Horizon labels are predicted based on concatenated visual segment features, geotemporal features and tabular morphological features (one label per segment).
    Soil images and data provided by: \cite{soil_fotos, soil_bgr}.}
    \label{fig:tasks}
\end{figure*}

As already pointed out in the introduction, to match the experts' approach when classifying soil horizons, we break down the general classification problem into three tasks, meant to fully describe the horizons:
Segmentation (Task 1), Morphological Feature Prediction (Task 2) and Horizon Classification (Task 3), see  \autoref{fig:tasks}).
\revisionchange{We use standard mathematical notations throughout the following subsections. Symbols written in bold, such as $\mathbf{z}$ and $\mathbf{x}$ denote vectors (either vectors of extracted features or vectors of predictions made within the model). A real-valued vector of dimension $n$ will be written as $\mathbf{x} \in \mathbb{R}^n$. Functions applied on such vectors are denoted by $f(\mathbf{x})$.}

\subsection{SoilNet}\label{subsec:soilnet}

\begin{figure*}[pos=h]
    \centering
    \includegraphics[width=\linewidth]{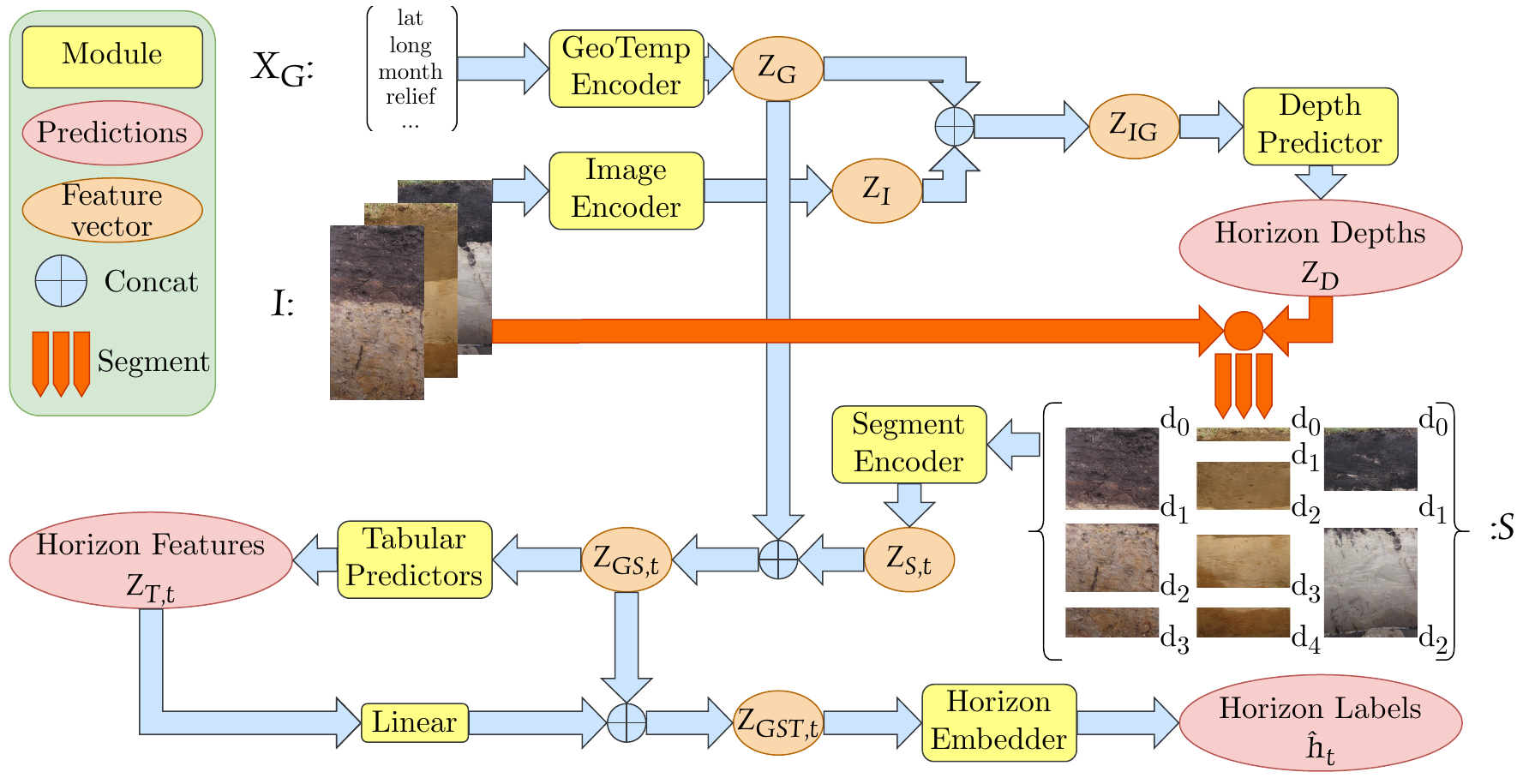}
    \caption{\textbf{SoilNet-Architecture.} An illustration of our proposed modularized multimodal multitask architecture for solving the three tasks (\textbf{SoilNet}). 
    Soil images and data provided by: \cite{soil_fotos, soil_bgr}.}
    \label{fig:soilnet}
\end{figure*}

The inner mechanisms of SoilNet are depicted in \autoref{fig:soilnet}. Our model processes images as the one shown in \autoref{fig:profile_example} and geotemporal metadata to sequentially predict depth markers, extract morphological (tabular) properties and classify soil horizons using a graph-based embedding approach. Our ablation studies in \autoref{sec:results} compare this to a straightforward cross-entropy loss. Below, we describe each key component of SoilNet in short. For more technical details regarding the implementation of the modules and the models, please consult our repository.

\paragraph{Image Encoder.} 

The image encoder extracts visual features from the soil profile images. Given an input RGB-image $I$, SoilNet applies a convolutional neural network (CNN) to extract visual features:  
\begin{equation}
\mathbf{z}_I = f_{\text{img}}(I) \in \mathbb{R}^{d_I}
\end{equation}
where $ f_{\text{img}} $ represents the CNN-based feature extractor and $ d_I $ is the dimensionality of the extracted feature vector. The soil profile images in the data exhibit significant variations in shape and resolution. Batch-processing requires all images to have the same dimensions. However, resizing the images to a dimension too low would lead to a distortion in the profile geometry and significant blurring of the very fine horizon features. We avoided this by padding the images to equal sizes. In the image encoder, we used a ResNet18 backbone~\citep{resnet} (initialized with ImageNet weights).
Instead of standard average pooling, we compute a masked average pooling to only consider features corresponding to the unpadded image. The output is afterwards linearly projected onto the $d_I$-dimensional latent space. We denote this module as Masked ResNet.


\paragraph{Geotemporal Encoder.}

Geospatial and temporal information plays an important role in determining soil characteristics. Geotemporal features are represented as a vector 
\begin{equation}
\mathbf{x}_G = [\text{lat}, \text{long}, \mathds{1}(\text{month}), \mathds{1}(\text{year}), \mathds{1}(\text{relief type}), ...] \in \mathbb{R}^{G}
\end{equation}
where \textit{lat, long} correspond to geographic coordinates and $\mathds{1}(\cdot)$ are one-hot encodings of the categorical features. With acquiring this kind of data, the idea is that auxiliary relief and climate characteristics can be derived from the location and timestamp stored when taking the photo of the soil sample (for instance, on the surveyor's photo camera). 
The vector $x_G$ containing the geotemporal data is then processed by $f_{\text{geo}}$ - a fully connected neural network (MLP)  
\begin{equation}
\mathbf{z}_G = f_{\text{geo}}(\mathbf{x}_G) \in \mathbb{R}^{d_G}
\end{equation}
producing an encoded representation of geotemporal properties.  

\paragraph{Depth Predictor.} 

To segment the soil profile into horizons, we sequentially predict depth markers. Given the concatenated feature representation  
\begin{equation}
\mathbf{z}_{IG} = [\mathbf{z}_I \, ; \, \mathbf{z}_G] \in \mathbb{R}^{d_I+d_G}
\end{equation}
where $ [ \cdot \, ; \, \cdot] $ denotes concatenation along the feature axis, we use an LSTM module~\citep{lstm} preceded and succeeded by a linear layer to generate a sequence of depth markers:  
\begin{equation}
\mathbf{z}_D = [d_1, d_2, \dots, d_D], \quad d_t = f_{\text{depth}}(\mathbf{h}_{t-1}, \mathbf{z}_{IG}), \quad \forall t = 1, ..., D
\end{equation}
where $f_{\text{depth}}$ is the LSTM-based depth predictor, $\mathbf{h}_{t-1}$ is the hidden state at step $t-1$ and $D$ is the number of predicted horizons. Every $d_t \in (0, 1]$ is a proposal for the lower boundary of the horizon at step $t$. Depth markers in the dataset, originally given in centimeters $(0-100)$, were converted to meters. For example, $d_1 = 0.3$ indicates that the topmost horizon ends at $-0.3$ meters. We emphasize that soil samples can have a different number of horizons, so the vector $\mathbf{z}_D$ may have variable lengths for different inputs $\mathbf{z}_{IG}$. We decided to pad the output $\mathbf{z}_D$, so that $D$ is a fixed maximal sequence length for every sample. Concretely, once $d_t$ is within a margin of $\epsilon = 0.01$ meters to the value of a \textit{stop token} $s$ (here, 1 meter depth), the marker is rounded to $s$.

\revisionchange{We note that we experimented with a Cross Attention Transformer module instead of an LSTM-based one for the Depth Predictor, too. As shown in \autoref{sec:results}, the LSTM module outperforms the Cross Attention alternative. For this reason, only the LSTM-based Depth Predictor described here was made an integral part of the final SoilNet model.}

\paragraph{Segment Encoder.}

Once depth markers are predicted, they are used to crop the original profile images down to the proposed horizon segments.
Each segment $ [d_{t-1}, d_t] $ defines a horizontal stripe of the image (see \autoref{fig:profile_example}), which is processed by a second CNN-based encoder:  
\begin{equation}
\mathbf{z}_{S, t} = f_{\text{seg}}(I_{[d_{t-1}, d_t]}) \in \mathbb{R}^{d_S}
\end{equation}
where $f_{\text{seg}}$ represents the CNN backbone and $I_{[d_{t-1}, d_t]}$ is the cropped image region corresponding to the predicted soil horizon. For $t=1, d_0$ is set to 0, representing the zero marker on the soil surface. Similar to the image encoder, we need to handle segments of variable sizes. We experimented with two patch-based strategies for $f_{\text{seg}}$:
\begin{enumerate}
    \item \textbf{Grid-based Patching (PatchCNN)}: Segments are resized to $512 \times 1024$ pixels (a higher resolution than the standard 224-pixel requirement of standard visual CNNs). A $1 \times 2$ symmetrical grid is applied to extract patches of a fixed size $512 \times 512$, which are then processed by a custom 5-layer CNN.
    \item \textbf{Random Patching with ResNet}: From the segments cropped out of the original images, we randomly extract 48 patches of size $224 \times 224$. The patches are then processed by a standard ResNet18 backbone. 
\end{enumerate}
Both approaches average the feature vectors for all patches into one single feature vector for each segment. 
In \autoref{sec:results}, we compare the performance of the two approaches.

\paragraph{Tabular Predictors.}

At this stage, SoilNet predicts the morphological tabular features for each segmented region. For a given segment $ [d_{t-1}, d_t] $, the following vector is formed:
\begin{equation}
\mathbf{z}_{GS, t} = [\mathbf{z}_{G} \, ; \, \mathbf{z}_{S, t}] \in \mathbb{R}^{d_G+d_S} .
\end{equation}
This concatenated vector is processed by a set of tabular predictors of a similar LSTM-based architecture as the depth predictor's. The output is the set of tabular properties:  
\begin{equation}
\mathbf{z}_{T,t}^{i} = f_{\text{tab}}^{i}(\mathbf{z}_{GS, t}) \in \mathbb{R}^{m_i}, \quad \forall i = 1, ..., M ,
\end{equation}
where $ M $ is the number of morphological attributes e.g. soil color, humus class. The dimension $m_i$ of the $i$-th tabular feature vector is 1 for numerical features (like the percentage of coarse fragments) or the number of unique classes for categorical features (such as the humus content). $ f_{\text{tab}}^{i} $ is, hence, the module responsible for predicting the $i$-th tabular feature in the form $ \mathbf{z}_{T,t}^{i} $ for the horizon at step $t$. We denote by $\mathbf{z}_{T,t}$ the concatenation of all the outputs of the tabular predictors:
\begin{equation}
\mathbf{z}_{T,t} = [\mathbf{z}_{T,t}^{1} \, ; \, ... \, ; \, \mathbf{z}_{T,t}^{M}] \in \mathbb{R}^{m},
\end{equation}
where $m = \sum_i^M m_i$.

\paragraph{Horizon Embedder.}

Finally, soil horizons are classified using a graph-based embedding approach, which we extended from the algorithm in~\cite{embeddings_jena}. Each possible target horizon label $h_i$ (given as string or class index) is represented by an embedding vector
\begin{equation}
\varphi(h_i) \in \mathbb{R}^{d_E}, \forall i = 1, ..., N
\end{equation}
computed from a graph of horizon relationships, where edges represent hierarchical and compositional dependencies (see \autoref{fig:graph}). Note that $N \geq d_E$ since the embeddings of mixture labels are linear combinations of independent non-mixture embeddings. The horizon embedder $f_{\text{hor}}$ takes as input the final concatenated vector
\begin{equation}
\mathbf{z}_{GST,t} = [\mathbf{z}_{GS,t} \, ; \, \text{Lin}(\mathbf{z}_{T,t})] \in \mathbb{R}^{d_G+d_S+m_{\text{lin}}},
\end{equation}
where $\text{Lin}(\mathbf{z}_{T,t}) \in \mathbb{R}^{m_{\text{lin}}}$ is the linear projection of the tabular predictions $\mathbf{z}_{T,t}$ and 
\begin{equation}
    \mathbf{\hat{y}}_t =  f_{\text{hor}} (\mathbf{z}_{GST,t}) \in \mathbb{R}^{d_E}
\end{equation}
delivers the embedding prediction $\mathbf{\hat{y}}_t$ for the horizon candidate at step $t$.
The predicted horizon class $\hat{h}_t$ is inferred via a similarity score $s(\mathbf{\hat{y}}_t, \varphi(h_i))$ (for instance, cosine similarity) as 
\begin{equation}
\hat{h}_t = \argmax_{i} s(\mathbf{\hat{y}}_t, \varphi(h_i)).
\end{equation}
\autoref{sec:graph_emb} explains in detail how we computed the horizon embeddings and \autoref{sec:ex_emb} gives a numerical calculation example.

When training the whole SoilNet pipeline end-to-end, we optimized a weighted sum of all the individual losses:
\begin{equation}
    L_{total} = 10 L_{depth} + \frac{1}{10} L_{stones} + \sum_i^{M_{categ}} L_{categ_i} + 10 L_{horizon},
\end{equation}
where $M_{categ} = 5$ is the number of categorical morphological properties (soil texture, color, humus class, rooting and carbonates). The weights were derived from the individual loss scales when training the task solvers separately. Each loss can be scaled differently, if the user wishes to prioritize certain tasks, or the loss weights could also be learned as hyperparameters. Here, we assign as a default equal importance to all the tasks.

\subsection{Label Clustering}\label{sec:label_cluster}

The horizon classification problem (Task 3) is highly imbalanced, with some horizon labels being well-represented while others appear very rarely in the dataset (see \autoref{fig:horizon_histo}). Directly training a model on such an imbalanced label set leads to poor generalization for rare horizons and biased predictions towards the over-represented classes. To mitigate this, we employ a clustering approach that groups similar horizons while preserving the structure of the taxonomic hierarchy.   

First, we filter the set of horizon labels by retaining only those with more than 10 samples in the dataset. This ensures that every horizon class will have at least 2 samples in each of the training, validation and test set after the stratified splitting. Let $H$ be the full set of horizon labels. We define the filtered set of representative labels as $V = \{ h_1, h_2, \dots, h_N \} \subset H$ (which will also be the set of terminal nodes in the taxonomic graph described in \autoref{fig:graph} and \autoref{sec:graph_emb}). These selected labels form the classification targets for Task 3.  

Rare horizon labels $h_r \in H \setminus V$ with a maximum of 10 samples in the dataset are assigned to the closest label in $V$ using a string similarity measure. Specifically, we use the \textit{Levenshtein distance} $ d_L(h_r, h) $, which quantifies the number of character edits (insertions, deletions, substitutions) needed to transform one label into another~\citep{Turk_jellyfish_2023}. The assignment is performed as  
\begin{equation}
    h_r \mapsto \arg\min\limits_{h \in V} d_L(h_r, h)
\end{equation}
After applying the Levenshtein mappings, we needed to carry out a final rule-based correction step, since some of the mappings were not pedologically plausible. We implemented these rules using domain knowledge.

\subsection{Graph Embeddings}\label{sec:graph_emb}

\begin{figure*}[pos=h]
    \centering
    \includegraphics[width=1\linewidth]{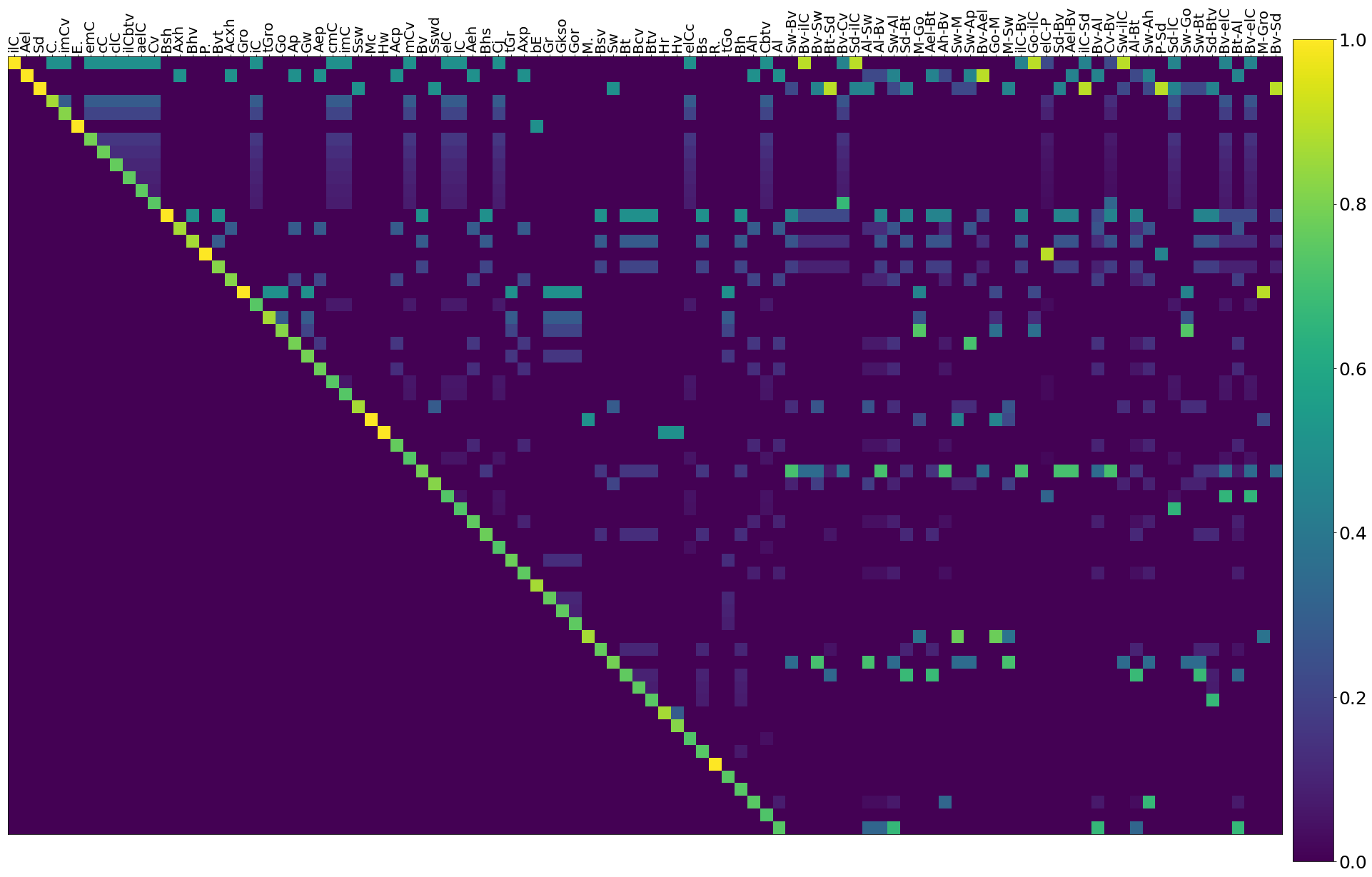}
    \caption{\textbf{Heatmap visualizing the horizon embeddings.} The matrix has dimension $(d_E \times N) = (61 \times 99)$. The left $61 \times 61$ block is diagonal and contains the embeddings of the non-mixture labels. The right $61 \times 38$ block contains the embeddings of the mixture labels.}
    \label{fig:emb_matrix}
\end{figure*}

Soil horizons exhibit complex relationships that cannot be adequately captured by a simple tree-based taxonomy. Instead, they form a \textit{graph-structured taxonomy}, where certain horizons have multiple parent nodes due to transitional characteristics (see \autoref{fig:graph}). To incorporate this structure into the model, we construct \textit{hierarchical embeddings} according to the method described in~\cite{embeddings_jena} and introduce an additional step to compute representations for complex horizons with multiple parent nodes.

Let the set of soil horizon labels be represented as a directed acyclic graph $ G = (V, \varepsilon) $, where $ V $ is the set of all horizon classes and $ \varepsilon \subset V \times V $ defines parent-child relationships\footnote{We use here the notation $\varepsilon$ to denote the edges referred to as $E$ in~\cite{embeddings_jena}, to avoid confusion with the symbol $E$ in $d_E$ we already introduced in \autoref{subsec:soilnet}.}. Furthermore, let $V^-$ be the subset of non-mixture horizon labels and $V^+$ the complement with mixture labels, so $V^- \cup V^+ = V$ and $V^- \cap V^+ = \emptyset$. The semantic similarity between two horizon labels $ h_i, h_j \in V^- $ is computed based on the height of their lowest common ancestor (LCA) in the hierarchy:  
\begin{equation}\label{eq:lca}
d_G(h_i, h_j) = \frac{\text{height}(\text{LCA}(h_i, h_j))}{\max\limits_{h \in V^-} \text{height}(h)}
\end{equation}
which is then converted into a similarity score:  
\begin{equation}\label{eq:sim}
s_G(h_i, h_j) = 1 - d_G(h_i, h_j).
\end{equation}
Following~\cite{embeddings_jena}, each non-complex horizon label $ h_i $ is embedded into a vector space such that their dot products reflect these similarity scores:  
\begin{equation}
\varphi(h_i)^\top \varphi(h_j) = s_G(h_i, h_j).
\end{equation}
To ensure all embeddings lie on a unit hypersphere, they are normalized as  
\begin{equation}
\|\varphi(h_i)\|_2 = 1.
\end{equation}
Embeddings are computed iteratively, starting from an abstract root node and adding new nodes while maintaining the required pairwise similarities. Please refer to \autoref{sec:ex_emb} for an example of computing several embeddings this way.

Complex horizons, which inherit properties from not one, but two parent nodes, are represented as a normalized linear combination of the embeddings of both their parents. Emphasizing the second symbol more, we obtained $\varphi(h_{\text{mix}})$ by normalizing
\begin{equation}\label{eq:eq_mixture}
\frac{1}{3} \varphi(h_{\text{parent}_1}) + \frac{2}{3} \varphi(h_{\text{parent}_2}) \quad \forall h_{\text{mix}} \in V^+ .
\end{equation}
This ensures that mixture horizons are positioned meaningfully within the embedding space, preserving their semantic relationships to both parent horizons. Instead of fixing the weights, one could learn them as hyperparameters. Following the German approach of horizon definitions~\citep{kartieranleitung5}, the correct classification of the second label of a complex horizon is more important, so we concluded that weighting $\varphi(h_{\text{parent}_2})$ twice as much as $\varphi(h_{\text{parent}_1})$ serves our purposes well enough. \autoref{fig:emb_matrix} displays the $d_E \times N$ embedding matrix as a color map. The first $d_E \times d_E$ block contains the independent non-mixture embeddings, while the remaining $d_E \times (N-d_E)$ block lists the mixture embeddings.

\subsection{Teacher Forcing}\label{subsec:teacher_forcing}

We also investigate the impact of different \textit{teacher forcing} strategies during training, inspired by similar standard approaches from sequence-to-sequence modeling~\citep{teacherforcing1}. Teacher forcing refers to the practice of providing the model with the ground-truth outputs from previous steps during training, rather than using its own predictions. This can help stabilize learning by preventing the accumulation of early prediction errors and encouraging faster convergence. However, if exaggerated, it can also make the model overly reliant on correct previous inputs, leading to poor performance at inference time. In our implementation, we apply teacher forcing within SoilNet at the stages where the inputs for learning Task 2 and Task 3 are generated, allowing us to control how much ground-truth information is exposed during sequential prediction. 

We experiment with two teacher forcing paradigms:

\begin{enumerate}
    \item \textbf{Full teacher forcing:} The model is trained with the ground-truth depth markers and tabular horizon features as inputs for all training samples. This means that the model is not exposed to any of its own predictions during training.
    \item \textbf{Linearly decreasing teacher forcing:} The model is trained with the ground-truth depth markers and tabular horizon features as inputs for the first five epochs. With each progressing epoch of these five, the teacher-forcing rate decreases, until the model is trained solely with its own predictions as inputs.
\end{enumerate}

\begin{revisionsection}
\subsection{Evaluation}\label{sec:evaluation}
We evaluate all models against the ground-truth annotations in a standard cross-validation setting. The data set was randomly split in three parts for training the models (training set, 60\% of the data), for model selection and optimizing hyperparameters (validation set, 20\% of the data) and a test set (20\% of the data) exclusively used for evaluation of the models' generalization performance. The hyperparameters can be found in \autoref{sec:appendix_exp_details}. For each of the tasks (segmentation, regression, classification) we used appropriate task-specific metrics described in detail below. 

\paragraph{Segmentation (Task 1): 1D-Intersection-over-Union (IoU).}
For horizon segmentation, we evaluate predicted depth marker sequences using a one-dimensional Intersection-over-Union (1D-IoU). We wrote a simplified form of the standard Intersection over Union (IoU) used for evaluating object detection models. Predicted and ground-truth depth markers define an ordered set of depth intervals (horizon stripes). We compute the overlap-to-union ratio for corresponding horizontal horizon stripes, assigning a zero intersection to the disjoint intervals.

\paragraph{Regression (Task 2): Mean Squared Error (MSE).}
For the numerical tabular feature \textit{Stones} (coarse fragments), we report Mean Squared Error (MSE) between predicted and ground-truth values.

\paragraph{Classification (Task 2 and 3): Precision/Recall@k.}
For categorical target variables in the tabular prediction task and the horizon prediction task, we report accuracy as well as precision and recall. Precision and recall were macro-averaged across classes to reduce the influence of frequent labels. We additionally report precision@k and recall@k, where a prediction is counted as correct if the ground-truth class is among the top-k ranked classes for a sample. This provides a more nuanced assessment of the models' ability to rank relevant labels effectively.

\paragraph{Aggregated Accuracy over Main Symbols (Task 3).}
To better understand the nature of misclassifications made by the models and to account for the hierarchical taxonomy, an aggregated accuracy is computed over the 10 main horizon symbols that constitute the first layer of the hierarchical label graph. This aggregation allows us to distinguish between minor label deviations, where the predicted label still falls into the correct main symbol, and complete structural misclassifications, where the prediction misses the correct main symbol. In the case of mixture horizons, the second label is accounted as the main symbol (see \autoref{eq:eq_mixture}).

\subsection{Baselines}
To understand the performance of the proposed SoilNet model, we implement two different baseline approaches. The first one leverages off-the-shelf large language models (LLMs) to solve simplified versions of Tasks 1 and 3 in a zero-shot manner with in-context-learning. The second baseline chains together individual task solvers for each of the three tasks, without joint training or end-to-end optimization. Below, we describe both baselines in detail.

\subsubsection{Baseline 1: Horizon Segmentation and Classification with LLMs}
\label{sec:llms}
We evaluate two general-purpose vision–language models in a zero-shot setting on simplified versions of Tasks 1 and 3 (5 training images, 5 testing images, and 12 horizon labels): Gemini 2.0 Flash~\citep{gemini2025} and ChatGPT-4o mini~\citep{chatgpt4omini2025}, both accessed between 5.-11.05.2025. 

Our approach was as follows:
\begin{itemize}
    \item[1)] We first explained the tasks to the LLM.
    \item[2)] We uploaded five random soil profile images from our training set in the chat along with their ground truth lists of depth markers and corresponding horizon labels. Here, we prompted the LLM to analyze the images with the accompanying target information and update its memory accordingly. In total, the five training images contained 12 different horizon labels.
    \item[3)] We uploaded 5 new images from our validation set and asked the LLM to predict the depths and horizon labels on its own. We selected the five images also randomly, but took care that they only contain horizon labels that the LLM has already seen in the previous five examples.
    \item[4)] We computed the 1D-IoU and top-1 horizon accuracy for the LLMs' predictions. 
\end{itemize}

The results are displayed together with the SoilNet results in \autoref{tab:soilnet_results}, partition A.
%
%
The chat histories are publicly available at: 
\begin{itemize}
    \item Gemini Chat History (Public Share):~\cite{geminiChatHistory2025}
    \item ChatGPT Chat History (Public Share):~\cite{chatgptChatHistory2025}
\end{itemize}

\subsubsection{Baseline 2: Chained Individual Task Solvers}
\label{sec:chained_task_solvers}
To compare SoilNet against a straightforward modular alternative, we chain three separately trained task solvers without any joint optimization or teacher forcing. Each solver is trained on its own task using ground-truth inputs and later frozen. At inference time, the solvers are executed sequentially, so that downstream modules must rely on the predictions of the upstream ones:
\begin{enumerate}
    \item \textbf{Depth prediction (Task 1):} The best-performing Masked ResNet image encoder + LSTM depth predictor (DP\_MaskedResNet\_LSTM) is trained on ground-truth depth markers.
    \item \textbf{Tabular prediction (Task 2):} The ResNet-based segment encoder with an LSTM head (TP\_ResNet\_LSTM) is trained on true horizon stripes and predicts the six tabular features. During chained inference, it consumes the segments produced by the depth predictor (Task 1).
    \item \textbf{Horizon classification (Task 3):} Two variants reuse the ResNet segment encoder with LSTM heads, differing only in the loss: the graph-embedding cosine loss (Baseline\_ResNet\_LSTM\_Emb) or the cross-entropy loss (Baseline\_ResNet\_LSTM\_CE). Training uses ground-truth tabular features and segments. During the chained inference, it consumes the predicted segments (Task 1) and tabular features (Task 2) from the previous solvers.
\end{enumerate}
This baseline isolates the effect of the pipeline coupling, where errors from depth prediction propagate into the estimation of the morphological tabular features and horizon classification. The resulting performance (see \autoref{tab:soilnet_results}, partition B) serves as a lower bound for modular approaches without end-to-end training.

\subsection{User Study}
\label{subsec:user_study_methods}
To assess the overall difficulty and benchmark SoilNet against human expertise, we conducted a user study with experienced soil surveyors. The expert annotators were provided only with the images, so that the annotators see the same information as SoilNet. This setup differs from the field annotators who curated the dataset and performed the assessment in the field, with access to additional contextual information beyond the image. In total, 15 users familiar with the task of soil horizon description participated in the study. Each one annotated between 1 and 47 soil profile images, with an average of 7 annotations per user.
We presented each user with randomly sampled soil profile images from our test set, while prioritizing profiles with existing annotations. Once a profile image reached 3 annotations, it was removed from the sampling pool. We further ensured that each user was only shown profiles images they had not annotated before by assigning anonymized annotator IDs. This approach allows us to quantify inter-annotator agreement and annotation variance per profile.

The user study involved a custom-developed web frontend, allowing the annotators to draw segment borders (Task 1) and annotate segments with horizon labels (Task 3), following the hierarchical taxonomy. The interface enforced the same taxonomy used throughout the paper. Screenshots of the frontend are provided in the appendix (\autoref{sec:user_study_screenshots}). We evaluate the annotations against the test set ground truth using the same metrics as SoilNet (see \autoref{sec:evaluation}): 1D-IoU for segmentation, plus top-1 accuracy and aggregated accuracy for horizon classification.
The results and visualizations are presented in \autoref{subsec:user_study_results}.

\end{revisionsection}

\section{Results}\label{sec:results}

This section presents the main results of our experiments
and is divided into two subsections: \textbf{Individual Tasks} and \textbf{End-to-End Pipeline}. 
The Individual Tasks denote the three tasks shown in \autoref{fig:tasks} solved disjointly (one separate model for each task),
while the End-to-End Pipeline refers to
the three tasks approached in a joint training and evaluation setting, as depicted in the SoilNet architecture in \autoref{fig:soilnet}. 
\revisionchange{We evaluate the models on the test set using the metrics described in \autoref{sec:evaluation}.}
The hyperparameters and compute resources are also provided in \autoref{sec:appendix_exp_details}.

\subsection{Individual Tasks}
\label{subsec:individual_tasks}

In the Individual Tasks stage, we train disconnected task solvers for the three proposed tasks. 
Concretely, the Individual Depth Predictor only solves Task 1 using full images and geotemporal features as input. The Individual Tabular Predictor only solves Task 2 using visual features from the horizon stripes (segmented with the ground truth depth markers) and geotemporal features. Finally, the Individual Horizon Predictor only solves Task 3 using features combined from the true horizon segments, geotemporal data and the true horizon tabular data. 
The insights gained from training the task solvers separately were subsequently used to construct the end-to-end SoilNet architecture. Additionally, 
the performance of the individual task solvers is meant
to estimate an upper bound for the end-to-end solver (SoilNet).
In the following, we will shortly discuss these insights and present the evaluating metrics for the different configurations. The best individual task solvers trained with the ground truth data will also be displayed in \autoref{tab:soilnet_results}. 

\paragraph{Task 1: Individual Depth Prediction.}
We first compared the performance of two depth predictor configurations, which differ in the depth prediction module itself: one LSTM-based, one Cross-Attention-based. In both cases, the visual features are extracted via the Masked ResNet Image Encoder described in \autoref{subsec:soilnet}.
The best Task 1 solver achieves an IoU of 51.74\%. For comparison, an average 'random chance' depth predictor\footnote{The average depth predictor simply responds with the average ground truth vector of depth markers computed on our training set.} achieves an IoU of 44.06\%.
\autoref{tab:simple_depth} shows the metrics for the two configurations of the individual depth predictor.

\begin{table*}[pos=h]
    \small 
      \caption{\textbf{LSTM-based depth predictor outperforms Cross-Attention module.} 
      The 1D-IoU scores are computed on the test set and given in percents (\%).
      \\ DP = Depth Predictor, MaskedResNet = image encoder and \{LSTM, CrossAtt\} = depth predicting module. 
      }
      \label{tab:simple_depth}
      \centering
      \begin{tabular}{lc}
        \toprule
        Model name         & IoU (\%) \\
        \midrule
        Average prediction     & 44.06 \\
        DP\_MaskedResNet\_LSTM   & \textbf{51.74} \\
        DP\_MaskedResNet\_CrossAtt   & 50.40 \\
        \bottomrule
      \end{tabular}
\end{table*}

\paragraph{Task 2: Individual Tabular Prediction.}\label{subsubsec:task2}
Secondly, we trained two configurations of Task 2 solvers by swapping the visual encoders for the true horizon segments: the custom PatchCNN and the  ResNet-based one, see \textit{Segment Encoder} in \autoref{subsec:soilnet}.
A ResNet-based segment encoder outperforms the custom PatchCNN encoder. The best Task 2 solver achieves an average accuracy of 47.59\% for the categorical features and a mean squared error (MSE) of 1.30 for the numerical feature (the percentage of stones).
\autoref{tab:aggregated_metrics_summary} shows the metrics for the two configurations of the individual tabular predictor. Both configurations use LSTM as tabular prediction modules for the six tabular horizon features. For better readability, we show the aggregated metrics for the categorical features in \autoref{tab:aggregated_metrics_summary}. The full metrics tables for the individual horizon-specific tabular predictions are shown in \autoref{sec:appendix_simple_tasks}, see \autoref{tab:simple_tabular_full}. \revisionchange{For a more detailed explanation of how precision@3 and recall@3 are computed, please consult \autoref{sec:evaluation}.}

\begin{table*}[pos=h]
    \small
        \centering
        \caption{\textbf{ResNet-based tabular predictor outperforms the custom PatchCNN-based tabular predictor.} 
        All metrics are computed on the test set and given in percents (\%), except MSE. The classification metrics are averaged over the six categorical features. See \autoref{tab:simple_tabular_full} for the full table.
        \\ TP = Tabular Predictor, \{PatchCNN, ResNet\} = segment encoder, LSTM = tabular prediction module, Acc. = Accuracy, Prec. = Precision, Rec. = Recall.}
        \label{tab:aggregated_metrics_summary}
            \begin{tabular}{l|c|ccccccc}
                \toprule
                & \multicolumn{1}{c|}{Stones} & \multicolumn{7}{c}{Categorical Features (Mean) - Metrics in \%} \\
                \cmidrule(r){2-2} \cmidrule(l){3-9}
                 & MSE & Acc. & F1 & Prec. & Rec. & Acc.@3 & Prec.@3 & Rec.@3 \\
                Model name &  &  &  &  &  &  &  &  \\
                \midrule
                TP\_PatchCNN\_LSTM & 6.11 & 44.53 & 20.22 & 27.47 & 21.39 & 74.67 & 61.68 & 47.52 \\
                TP\_ResNet\_LSTM & \textbf{1.30} & \textbf{47.59} & \textbf{24.15} & \textbf{29.44} & \textbf{25.52} & \textbf{77.92} & \textbf{68.01} & \textbf{54.14} \\
                \bottomrule
            \end{tabular}
\end{table*}


\paragraph{Task 3: Individual Horizon Prediction.}\label{subsubsec:task3}
Thirdly, we trained Task 3 solvers under the assumption that both the ground truth depth markers and the ground truth horizon-specific tabular features are provided as input. We evaluated four configurations: two visual segment encoders (PatchCNN vs ResNet-based) and two loss functions (cosine embedding loss - as described in \textit{Horizon Embedder} in \autoref{subsec:soilnet} - vs cross-entropy loss).
The main results are presented in \autoref{tab:simple_horizon}. The results highlight 
the difficulty of Task 3, particularly in recognizing horizons from the long tail of the label distribution (very low F1 scores). 
On the other hand, the top-5 metrics
indicate that the correct horizon label is frequently among the model's top-ranked predictions.
Similarly, the aggregated main symbol accuracies reveal that models are often able to correctly identify the primary symbol of a horizon.
This effect is especially pronounced in configurations trained with cosine embedding loss, suggesting that the learned label embeddings effectively model
both intra-class similarity and inter-class variability, leading to more geologically coherent predictions (see \autoref{sec:appendix_confusion_matrices}, where we present corresponding confusion matrices). \revisionchange{For a more detailed explanation of how precision@3 and recall@3 are computed, please consult \autoref{sec:evaluation}.}


\begin{table*}[pos=h]
    \small 
      \caption{\textbf{Graph embeddings lead to a higher aggregated accuracy over the main symbol than traditional one-hot encodings.} 
      All metrics are computed on the test set and given in percents (\%).
      \\ HP = Horizon Predictor, \{PatchCNN, ResNet\} = segment encoder, LSTM = horizon embedding/classification module, Emb = embedding loss, CE = cross entropy, Acc. = Accuracy, Prec. = Precision, Rec. = Recall, Agg. Acc. = Aggregated Accuracy over main symbols.}
      \label{tab:simple_horizon}
      \centering
      \begin{tabular}{lcccccc}
        \toprule
        Model name & Acc. & F1 & Acc.@5 & Prec.@5 & Rec.@5  & Agg. Acc. \\
        \midrule
        HP\_PatchCNN\_LSTM\_Emb & 47.62          & 14.23          & 71.68          & 55.04          & 47.21          & \textbf{78.99} \\
        HP\_ResNet\_LSTM\_Emb   & 47.75          & \textbf{14.69} & 72.27          & 50.56          & \textbf{49.01} & \textbf{78.99} \\
        HP\_PatchCNN\_LSTM\_CE  & \textbf{51.55} & 10.07          & \textbf{78.89} & 56.38          & 37.55          & 73.84          \\
        HP\_ResNet\_LSTM\_CE    & 48.14          & 11.13          & 78.88          & \textbf{59.00} & 44.03          & 69.41          \\
        \bottomrule
      \end{tabular}
\end{table*}

\subsection{End-to-end Pipeline: SoilNet}
With what we have learned from training the individual task solvers above, we have designed the SoilNet architecture - an end-to-end pipeline capable of solving all three tasks simultaneously - as shown in \autoref{fig:soilnet}.

\autoref{tab:soilnet_results} compares the results for various configurations of the SoilNet model (partitions C and D) with  baseline models (partitions A and B) and the individual task solvers' results (partition E). \autoref{tab:soilnet_results} (partition A) reports the \revisionchange{results of Baseline 1 (see also \autoref{sec:llms})}, where we conducted a zero-shot performance of two LLMs for simplified versions of Tasks 1 and 3.
The Baseline 2 models in \autoref{tab:soilnet_results} (partition B) are chained individual task solvers\revisionchange{, where} each solver consumes the outputs of the respective previous task solver \revisionchange{(see also \autoref{sec:chained_task_solvers})}. In partition E, we  report the best performing module configuration for individual task solvers with access to the ground truth of the respective previous task (for a more detailed overview, consult the previous \autoref{subsec:individual_tasks}). Note that since ground truth from previous tasks is not available in real world applications, we do not include these in the model comparison. 
Results in partitions B and E can be considered lower bounds (using naive chaining of Task solvers' outputs) and upper bounds (with access to ground truth of previous tasks) for each task. 

\begin{table*}[pos=h]
    \small 
      \caption{
      \textbf{
      SoilNet (D) outperforms baseline of zero-shot LLMs (A) and chained task solvers (B) in most tasks.} For context, we show the optimal performance for all tasks obtained by models with access to ground truth of respective previous tasks at inference (E), but not include them in comparison. All test set metrics are given in percents (\%), except for Stones (coarse fragments) MSE. 
      SN = SoilNet, DP = Depth Predictor, TP = Tabular Predictor, HP = Horizon Predictor, \{PatchCNN, ResNet\} = segment encoder, LSTM = horizon embedding/classification module, Emb = embedding loss, CE = cross entropy, Acc. = Accuracy, Agg. Acc. = Aggregated Accuracy over main symbols (see \autoref{sec:appendix_exp_details} for evaluation metrics and more experiment details).}
      \label{tab:soilnet_results}
      \centering
      \resizebox{\textwidth}{!}{
      \begin{tabular}{l | c | cc | cccccc}
        \toprule
        \multirow{3}{*}{Model name} & \multicolumn{1}{c|}{Depths} & \multicolumn{2}{c|}{Tabular} & \multicolumn{6}{c}{Horizon Label} \\
        \cmidrule(r){2-2} \cmidrule(lr){3-4} \cmidrule(l){5-10}
    
        & \multirow{2}{*}{IoU} & Stones & Categ. & \multirow{2}{*}{Acc.} & \multirow{2}{*}{F1} & Acc. & Prec. & Rec. & Agg. \\
    
        &                      & MSE    & Acc.   &                       &                     & @5   & @5   & @5   & Acc. \\
        \midrule
        \multicolumn{10}{c}{\textbf{A. \revisionchange{Baseline 1:} Zero-shot LLM on simplified Task 1 and 3 (5 images, 12 horizon labels, see \autoref{sec:llms})}} \\
        \midrule
        Gemini 2.0 Flash   & 41.09 & -             & -              & 40.0  & -              & -              & -              & -              & -              \\
        ChatGPT-4o mini   & 36.08          & -             & -              & 32.0           & -              & -              & -              & -              & -              \\
        \midrule
        \multicolumn{10}{c}{\textbf{B. \revisionchange{Baseline 2:} Individual task solvers (using predictions of previous tasks\revisionchange{, see \autoref{sec:chained_task_solvers}})}} \\
        \midrule
        Baseline\_ResNet\_LSTM\_Emb  & \textbf{51.74} & \textbf{1.58} & 46.22 & 26.40 & 0.85  & 33.54          & 15.44          & 9.91  & 45.27 \\
        Baseline\_ResNet\_LSTM\_CE   & \textbf{51.74} & 1.59          & 46.12          & 25.34          & 0.81           & 47.05 & 22.46 & 9.64           & 41.17          \\
        \midrule
        \multicolumn{10}{c}{\textbf{C. SoilNet (Full Teacher Forcing)}} \\
        \midrule
        SN\_PatchCNN\_LSTM\_Emb & 51.41 & 8.03           & 46.82          & 25.91          & 1.87  & 39.97          & 25.45          & 15.80 & 52.35          \\
        SN\_ResNet\_LSTM\_Emb   & 50.90          & 1.60  & 45.84          & 24.84          & 0.80           & 33.91          & 17.07          & 10.03          & 42.84          \\
        SN\_PatchCNN\_LSTM\_CE  & 49.53          & 8.89           & 47.72 & 29.85 & 1.62           & 56.43          & 21.34          & 13.04          & 53.58 \\
        SN\_ResNet\_LSTM\_CE    & 51.41 & 1.99           & 47.39          & 27.90          & 1.48           & 57.61 & 28.00 & 15.37          & 48.60          \\
        \midrule
        \multicolumn{10}{c}{\textbf{D. SoilNet (Linearly decreasing Teacher Forcing over 5 Epochs)}} \\
        \midrule
        SN\_PatchCNN\_LSTM\_Emb & 51.25          & 8.92           & 46.60          & 36.27          & 7.55           & 60.21          & 40.27          & 33.84          & \textbf{71.42} \\
        SN\_ResNet\_LSTM\_Emb   & 51.47 & 1.74  & 47.64          & 35.40          & 6.58           & 59.65          & 33.75          & 33.08          & 68.70          \\
        SN\_PatchCNN\_LSTM\_CE  & 49.52          & 12.38          & 46.47          & 43.99          & 7.20           & 72.02          & \textbf{50.33} & 30.33          & 68.61          \\
        SN\_ResNet\_LSTM\_CE    & 49.91          & 1.99           & \textbf{48.08} & \textbf{45.70} & \textbf{7.99}  & \textbf{76.25} & 49.85          & \textbf{35.03} & 69.88          \\
        \midrule
        \midrule
        \multicolumn{10}{c}{\textbf{E. Individual task solvers (using ground truth of previous task for inference, see \autoref{subsec:individual_tasks})}} \\
        \midrule
        DP\_MaskedResNet\_LSTM  & 51.74 & -              & -              & -              & -              & -              & -              & -              & -              \\
        TP\_ResNet\_LSTM        & -              & 1.30  & 47.59 & -              & -              & -              & -              & -              & -              \\
        HP\_ResNet\_LSTM\_Emb   & -              & -              & -              & 47.75          & 14.69 & 72.27          & 50.56          & 49.01 & 78.99 \\
        HP\_ResNet\_LSTM\_CE    & -              & -              & -              & 48.14 & 11.13          & 78.88 & 59.00 & 44.03          & 69.41          \\
        \bottomrule
      \end{tabular}
      }
\end{table*}

For SoilNet, we compare the impact of swapping two different visual segment encoders (PatchCNN vs ResNet-based) and loss functions (embedding-based vs cross-entropy). 
In partitions C and D the results for these four variants are presented, obtained using the two teacher forcing strategies introduced in \autoref{subsec:teacher_forcing}.
Between the two strategies, we observe that the linearly decreasing teacher forcing leads to much better results for the horizon label prediction metrics across all model configurations. This is likely due to the fact that, in the second teacher strategy, the models are gradually exposed to their own predictions, which helps them to learn to handle noise and uncertainty from their own predictions. The full teacher forcing, on the other hand, can lead to overfitting on the ground-truth inputs and poor generalization in inference. For depth prediction and tabular metrics, the performance differences are less pronounced, as the ground-truth inputs do not affect the models as much at these stages.
Partition D shows the SoilNet trained with the winning teacher strategy.

\begin{revisionsection}
\subsection{User Study Results}
\label{subsec:user_study_results}
In the user study (\autoref{subsec:user_study_methods}), we compare expert annotations with predictions from the SoilNet variant with the highest horizon-classification accuracy (SN\_ResNet\_LSTM\_CE, \autoref{tab:soilnet_results}, partition D). The variant uses ResNet-based encoders for the profile image and the horizon segments, LSTM heads for the task predictors and a cross-entropy loss for horizon labels. \autoref{fig:user_annotations} shows 2 example profiles with the ground truth as provided in the data, our model's predictions, and the annotations from 2 users for these images. In Sample 1 (Subfigures \ref{subfig:user_annotations_a}–\ref{subfig:user_annotations_d}), the users draw fewer boundaries than the model and assign different labels for several segments. In Sample 2 (Subfigures \ref{subfig:user_annotations_e}–\ref{subfig:user_annotations_h}), our SoilNet model misses a narrow B horizon that both users annotate. The summary metrics in Subfigure  \ref{subfig:user_annotations_i} report segmentation and horizon-classification scores for the model and the user annotations on the study images. Using our established evaluation metrics (\autoref{sec:evaluation}) of segmentation IoU, horizon-classification accuracy, and aggregated accuracy, we observe that SoilNet generally achieves a higher predictive performance than users when only images are provided.

\begin{figure*}[pos=h]
    \centering
    Sample 1\\
    \vspace*{-1em}
    \subfloat[Ground Truth]{\includegraphics[height=5cm]{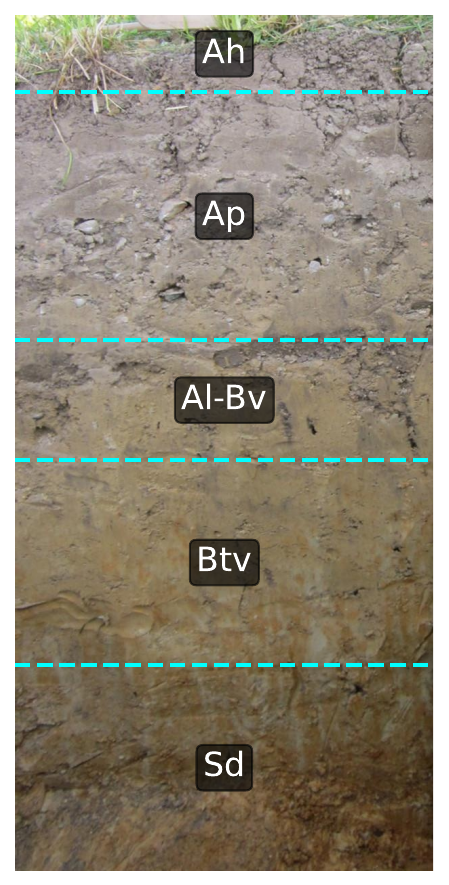}\label{subfig:user_annotations_a}}
    \hspace{0.05\textwidth} 
    \subfloat[SoilNet]{\includegraphics[height=5cm]{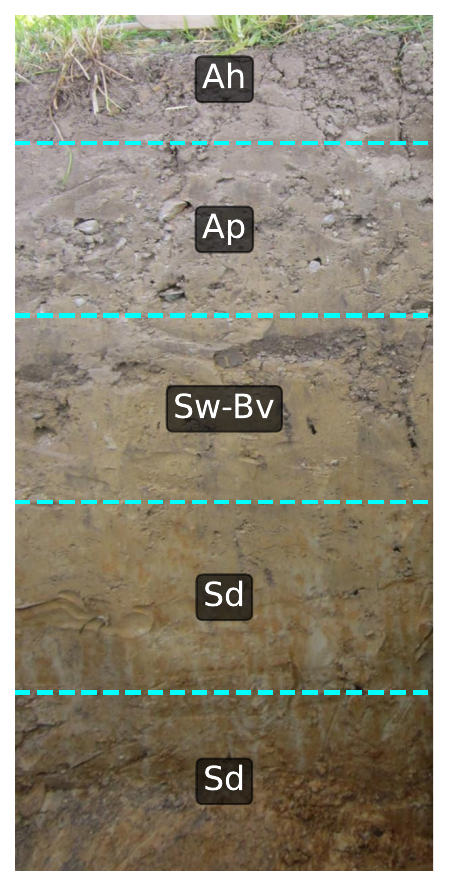}\label{subfig:user_annotations_b}}
    \hspace{0.05\textwidth} 
    \subfloat[User 1]{\includegraphics[height=5cm]{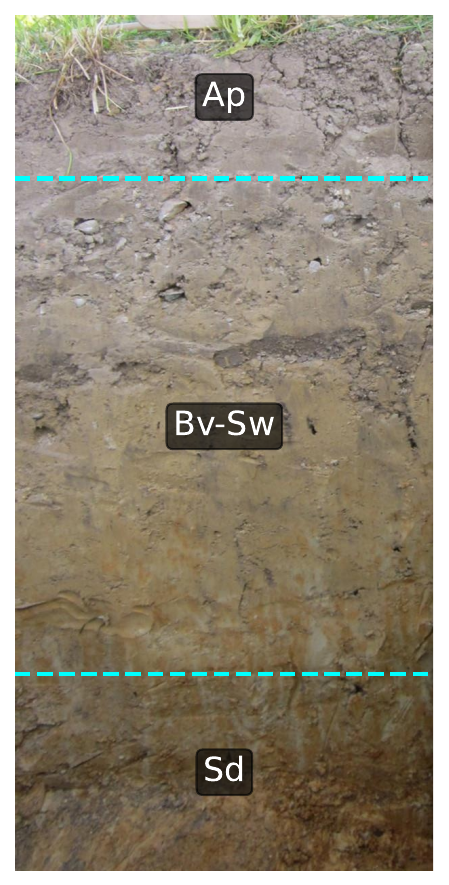}\label{subfig:user_annotations_c}}
    \hspace{0.05\textwidth} 
    \subfloat[User 2]{\includegraphics[height=5cm]{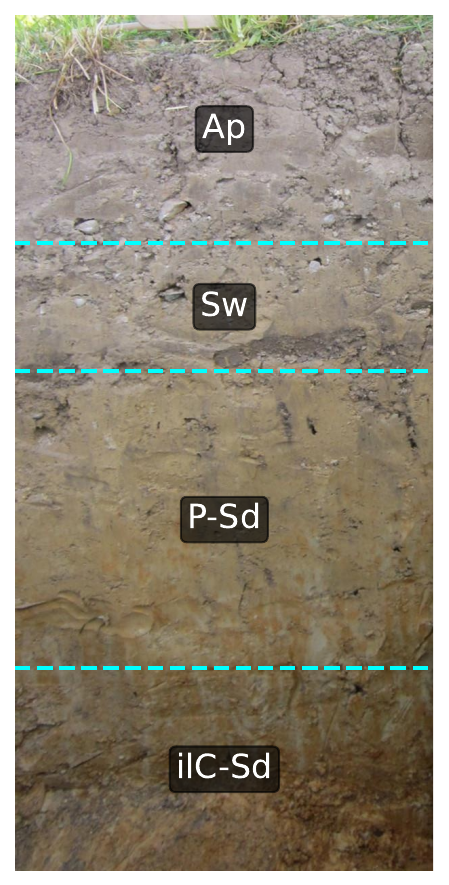}\label{subfig:user_annotations_d}}
    \vspace*{1em}
    
    Sample 2\\
    \vspace*{-1em}
    \subfloat[Ground Truth]{\includegraphics[height=5cm]{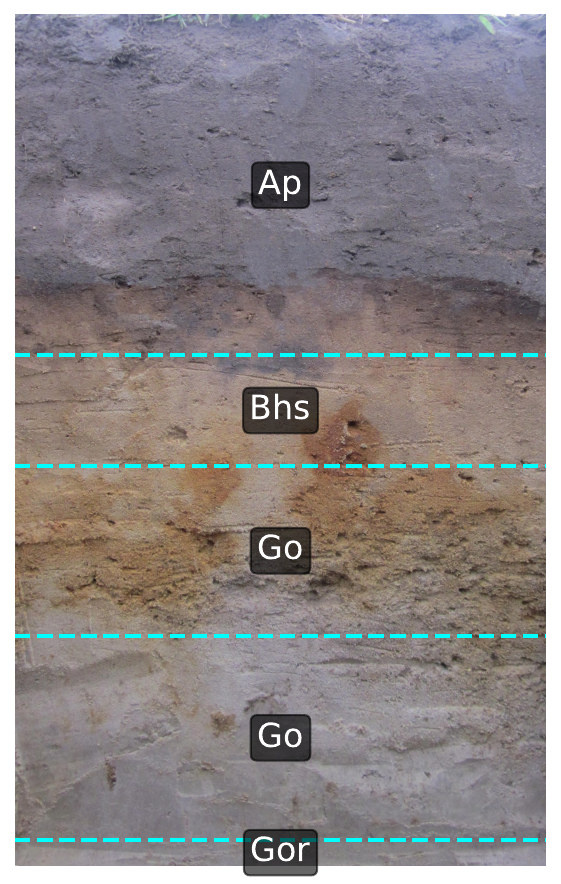}\label{subfig:user_annotations_e}}
    \hspace{0.05\textwidth} 
    \subfloat[SoilNet]{\includegraphics[height=5cm]{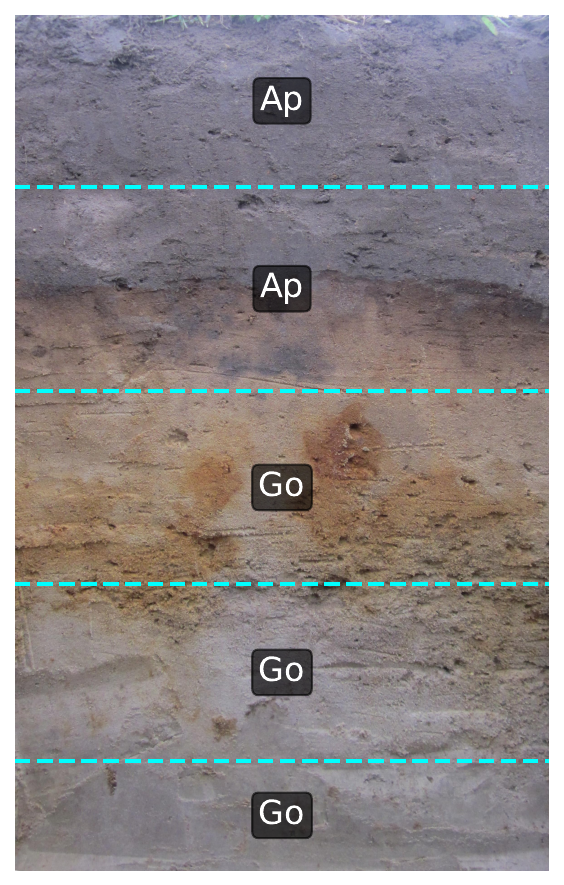}\label{subfig:user_annotations_f}}
    \hspace{0.05\textwidth} 
    \subfloat[User 1]{\includegraphics[height=5cm]{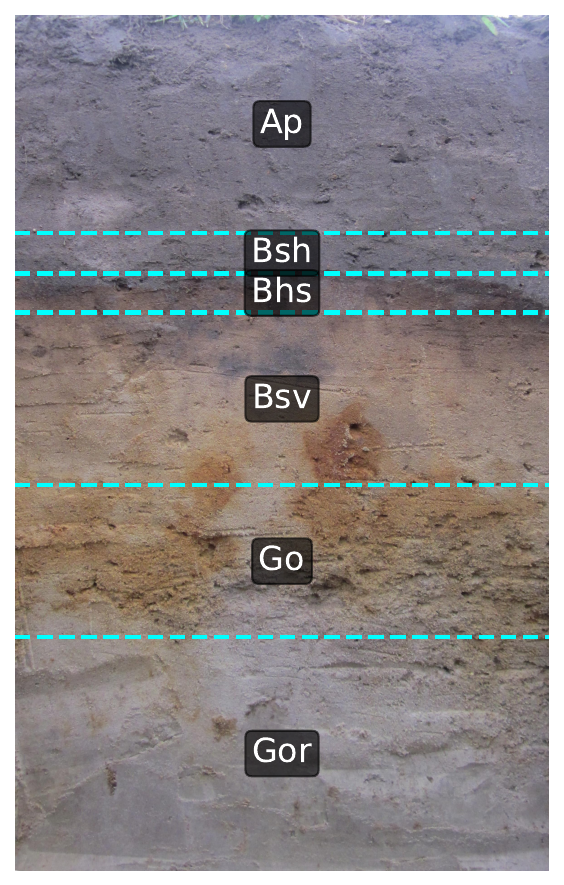}\label{subfig:user_annotations_g}}
    \hspace{0.05\textwidth} 
    \subfloat[User 2]{\includegraphics[height=5cm]{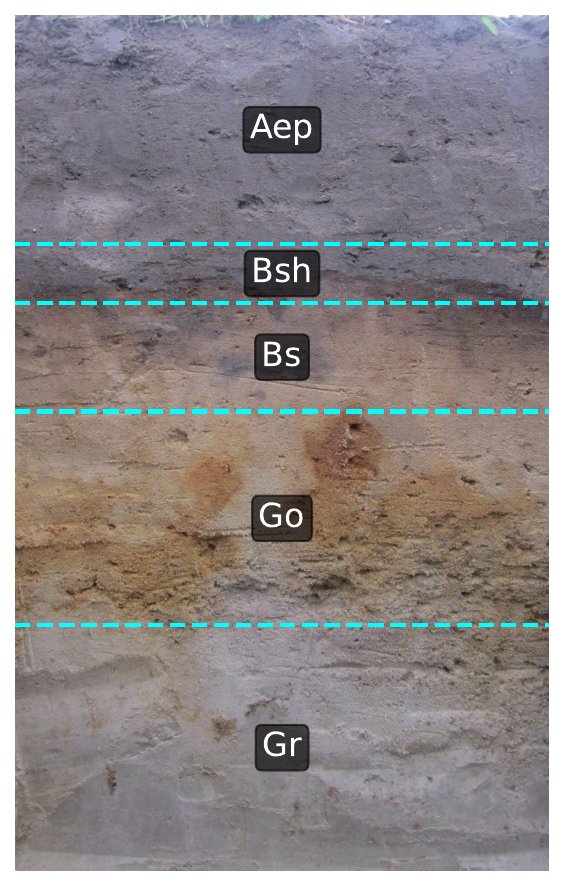}\label{subfig:user_annotations_h}}

    \subfloat[Comparison of Metrics]{\includegraphics[height=3.45cm]{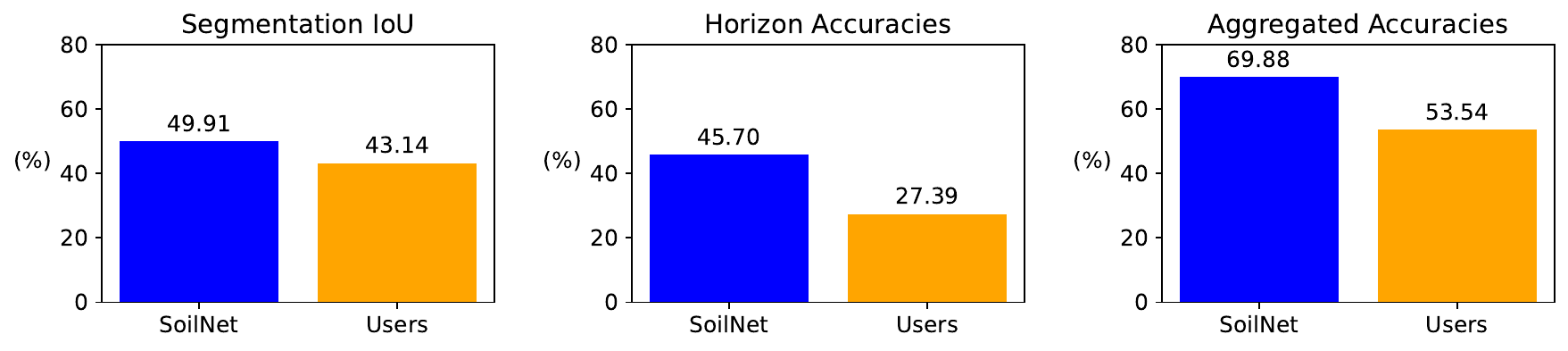}\label{subfig:user_annotations_i}}
    \caption{\revisionchange{\textbf{Illustration of task difficulty and predictive performance of SoilNet in comparison to human experts.} User study results indicate that for Task 1 (segmentation) and Task 3 (horizon classification), SoilNet can achieve predictive performances better than human experts (i). SoilNet performs on par with human experts and reproduces the ground truth data of the test set efficiently, as shown in a representative sample 1 (b). In sample 2 (f), however, SoilNet fails to detect the narrow B horizon, which is identified by human experts (g and h).}}
    \label{fig:user_annotations}
\end{figure*}
\end{revisionsection}

\section{Discussion}\label{sec_dis}
\revisionchange{The results highlight key differences between general-purpose and task-specific models, the benefits of end-to-end multitask learning, and the influence of intermediate predictions on overall performance. The following sections discuss these findings, assess the overall task difficulty, and outline practical implications, limitations, and directions for future work.}

\subsection{SoilNet is more accurate than baselines}
Our results highlight the importance of architectures specialized for tasks such as soil modeling. We find that zero-shot inference with in-context-learning using standard LLMs (\autoref{tab:soilnet_results}, partition A) yields poor performance for the complex soil horizon classification task, even on simplified versions of Task 1 and Task 3. Both LLMs evaluated achieve IoU scores below the random guessing baseline (\autoref{tab:simple_depth}) and top-1 horizon classification accuracy lower than several SoilNet variants trained on the full set of 99 horizon labels. These findings hint at the limitations of general-purpose LLMs in addressing specialized, complex domain-specific problems such as soil horizon classification.
    
Moreover, SoilNet outperforms the pipeline of chained individual task solvers (\autoref{tab:soilnet_results}, partitions B and D) across most tasks; for the regression task of predicting the percentage of stones (coarse fragments) and the depth marker segmentation, SoilNet achieves slightly lower but comparable results. This indicates that jointly optimizing all tasks in an end-to-end multitask setting yields higher predictive performance than independently trained modules coupled only at inference time.  
  
The individually trained task-specific solvers using ground truth information from the respective previous tasks (\autoref{tab:soilnet_results}, partition E) establish strong upper bounds for each task. Note that these may not be used for prediction in real applications, since they require ground truth information. SoilNet, without access to that ground truth information, often achieves similar predictive performance. 
The difference between these upper bounds using ground truth and SoilNet for horizon label metrics suggests that performance in the intermediate tasks (depth marker and tabular prediction) has a great impact on the final horizon classification performance.  
      
Within the SoilNet configurations (\autoref{tab:soilnet_results}, partition D), the variant using the ResNet-based segment encoder generally surpasses the PatchCNN alternative across most evaluation metrics. Regarding training objectives, the best performance in terms of aggregated main symbol accuracy is obtained using the \textit{embedding-based cosine loss}, indicating that semantically structured label representations provide a more effective learning signal than conventional categorical targets. The differences in the aggregated accuracy between the different SoilNet configurations are, however, not as pronounced as in the individual Task 3 solvers.

\subsection{Task Difficulty and Expert Assessment}
\revisionchange{The user study (\autoref{subsec:user_study_methods}, \autoref{subsec:user_study_results}) and \autoref{fig:user_annotations} highlights that horizon annotation from images alone remains challenging for both experts and SoilNet. For some examples (Sample 1,  Subfigures \ref{subfig:user_annotations_a}--\ref{subfig:user_annotations_d}), the task appears more difficult for users, while for others (Sample 2, Subfigures \ref{subfig:user_annotations_e}--\ref{subfig:user_annotations_h}) it is more difficult for our SoilNet model. The disagreement in horizon labels between the ground truth, SoilNet, and the expert annotators indicates that assigning horizon labels (Task 3) is the most challenging component when only faced with the profile image, due to the complexity of the underlying taxonomy. Nonetheless, our SoilNet model typically delivers pedologically sound predictions. For example, it never predicts \textit{A} horizons below \textit{B} or \textit{C} horizons and, in 86\% of the samples containing an \textit{Ap} horizon, this is predicted as topmost.}

\subsection{Implications and Use cases}
\revisionchange{Our evaluations highlight the potential of SoilNet for a number of use cases, such as (i) assessing the quality of incoming soil profile descriptions in data repositories or land and soil information systems, (ii) retrieving appropriate data for studies conducted within such systems, and (iii) identifying study sites in cases where tabular data are insufficient. The horizon labels of the German classification system used in this study encode morphological soil characteristics, which are also predicted as part of Task 2. Hence, the approach could also support the identification of such characteristics in the field as part of an expert system that guides soil scientists during profile description, for example by prompting further investigation of recognizable but not yet described soil features or by suggesting possible alternative genetic interpretations of those features.}

Beyond the specific use case addressed in this paper, the modular architecture we present - designed for multimodal classification with intra-sample sequential dependencies and a hierarchical label structure - may serve as a generalizable template for related problems in other domains. Similar settings arise in tasks such as: i) \textit{medical imaging}, where multimodal inputs e.g. scans and patient metadata must be segmented into anatomically meaningful regions with hierarchical diagnoses~\citep{med_img_metadata,bone_tumor}; ii) \textit{remote sensing}, where satellite imagery (combined multispectral and hyperspectral) is segmented into land cover zones that follow spatial sequences and taxonomies~\citep{sattelite_multi_hyper}; iii) \textit{document layout analysis}, where textual and visual cues jointly inform the sequential classification of structured document elements~\citep{doc_layout}.


\subsection{Limitations} 
Several simplifications and design choices introduce limitations in our SoilNet pipeline. First, we trained our model on a reduced set of horizon labels. While this was necessary to manage class imbalance, it also limits the model’s ability to generalize to the full complexity of the horizon taxonomy, where label distributions are significantly more skewed. 
\revisionchange{One consequence of this simplification and the associated class imbalance concerns buried horizons. Due to their very low occurrence in the dataset (approximately 1.5\% of horizons), the model is unlikely to have learned a robust representation of this horizon group. Moreover, during label mapping to ensure sufficient samples across training, validation, and test splits, horizons with an \textit{f}-prefix (buried horizons in the German labeling system) were remapped to more frequent classes based on symbol similarity, such that the explicit information about buried horizons may have been lost.}
Strategies such as data augmentation, class balanced losses or over-/under-sampling may help address this issue.
Second, in modeling mixture horizons, we assigned fixed weights of the parent embeddings to each mixture label (see \autoref{eq:eq_mixture} in \autoref{sec:graph_emb}). More nuanced strategies, such as learning mixture weights as trainable parameters or defining expert-informed weights per mixture, could better capture the semantic asymmetry and domain-specific knowledge involved in these transitions. Third, despite their semantic grounding, our graph-based label embeddings are outperformed on certain metrics by simpler cross-entropy-trained configurations of SoilNet, suggesting that the added structure does not always translate into performance gains across all tasks.

\subsection{Future Work} 
In future work, we plan to develop an application that integrates the SoilNet pipeline into an accessible interface for use in the field. Such a tool would allow users to capture a soil profile image, automatically retrieve geotemporal metadata and obtain real-time predictions for depth segmentation, morphological features and horizon classification. \revisionchange{The application could serve as an expert system guiding users through the overall tasks of profile description.} This would facilitate practical deployment for soil or environmental scientists, agronomists or even citizen scientists.
Additionally, it would be valuable to conduct a systematic evaluation of \revisionchange{experienced soil surveyors annotating soil profiles with the support of our SoilNet model. Such a study could analyze how model suggestions influence expert decision-making, including which predictions are accepted or revised}. This would not only offer insight into our model's practical usefulness, but may also reveal how automation can assist or complement professional interpretation.


\section{Conclusion}\label{sec_con}
In this work, we introduced SoilNet, a modular, multimodal, multitask framework for automated soil horizon classification, designed to mirror the structured reasoning process of human experts. \revisionchange{Empirical evaluations demonstrate that SoilNet can achieve performances better than both simple single task foundation models as well as high-capacity generic LLM systems. Results from a user study with a custom application for annotating soil horizons indicate that the predictive performance of SoilNet can be on par with, or better than, that of human experts when both are limited to profile images only. It should be noted, however, that human experts in field settings would very likely outperform SoilNet due to their access to additional contextual information beyond the profile image.} Beyond its technical contributions, our research also seeks to provide a means for updating soil data, soil data quality improvements and for soil data acquisition. As soil degradation continues to threaten food security and biodiversity, the development of modern, scalable, domain-adapted AI solutions becomes increasingly urgent. Our work demonstrates how advances in Machine Learning, when carefully aligned with domain knowledge, can contribute to efficient soil monitoring and enhance scientific understanding.

\printcredits

\section*{Data Availablility}
We are currently in contact with the Thünen institute to prepare the publication of the dataset used. Until then, we refer the readers to a limited dataset already published in~\cite{soil_fotos}.

\section*{Declaration of competing interests}
The authors declare that they have no known competing financial interests or personal relationships that could have appeared to influence the work reported in this paper.

\section*{Acknowledgements}
This research was supported by the German Research Foundation (DFG) - Project number: 528483508 - FIP 12.

\section*{Declaration of generative AI and AI-assisted technologies in the manuscript preparation process}
During the preparation of this work the authors used Google Gemini and ChatGPT in order to perform minor edits. After using these tools/services, the authors reviewed and edited the content as needed and take full responsibility for the content of the published article.

%
%

\bibliographystyle{cas-model2-names}
\bibliography{references}

\begin{thebibliography}{58}
\expandafter\ifx\csname natexlab\endcsname\relax\def\natexlab#1{#1}\fi
\providecommand{\url}[1]{\texttt{#1}}
\providecommand{\href}[2]{#2}
\providecommand{\path}[1]{#1}
\providecommand{\DOIprefix}{doi:}
\providecommand{\ArXivprefix}{arXiv:}
\providecommand{\URLprefix}{URL: }
\providecommand{\Pubmedprefix}{pmid:}
\providecommand{\doi}[1]{\href{http://dx.doi.org/#1}{\path{#1}}}
\providecommand{\Pubmed}[1]{\href{pmid:#1}{\path{#1}}}
\providecommand{\bibinfo}[2]{#2}
\ifx\xfnm\relax \def\xfnm[#1]{\unskip,\space#1}\fi
\bibitem[{{Ad-hoc-AG Boden}(2005)}]{kartieranleitung5}
\bibinfo{author}{{Ad-hoc-AG Boden}}, \bibinfo{year}{2005}.
\newblock \bibinfo{title}{Bodenkundliche Kartieranleitung, 5. Auflage [KA5 - Soil Survey Guidelines, 5th ed.]}.
\newblock \bibinfo{publisher}{Schweizerbart Science Publishers}, \bibinfo{address}{Hannover, Germany}.
\newblock \URLprefix \url{https://www.schweizerbart.de/publications/detail/isbn/3510959205}.
\bibitem[{Adeniyi et~al.(2024)Adeniyi, Brenning and Maerker}]{soils_lombardia}
\bibinfo{author}{Adeniyi, O.D.}, \bibinfo{author}{Brenning, A.}, \bibinfo{author}{Maerker, M.}, \bibinfo{year}{2024}.
\newblock \bibinfo{title}{Spatial prediction of soil organic carbon: Combining machine learning with residual kriging in an agricultural lowland area (lombardy region, italy)}.
\newblock \bibinfo{journal}{Geoderma} \bibinfo{volume}{448}, \bibinfo{pages}{116953}.
\newblock \URLprefix \url{https://www.sciencedirect.com/science/article/pii/S0016706124001824}, \DOIprefix\doi{https://doi.org/10.1016/j.geoderma.2024.116953}.
\bibitem[{Alayrac et~al.(2022)Alayrac, Donahue, Luc, Miech, Barr, Hasson, Lenc, Mensch, Millican, Reynolds, Ring, Rutherford, Cabi, Han, Gong, Samangooei, Monteiro, Menick, Borgeaud, Brock, Nematzadeh, Sharifzadeh, Binkowski, Barreira, Vinyals, Zisserman and Simonyan}]{flamingo}
\bibinfo{author}{Alayrac, J.B.}, \bibinfo{author}{Donahue, J.}, \bibinfo{author}{Luc, P.}, \bibinfo{author}{Miech, A.}, \bibinfo{author}{Barr, I.}, \bibinfo{author}{Hasson, Y.}, \bibinfo{author}{Lenc, K.}, \bibinfo{author}{Mensch, A.}, \bibinfo{author}{Millican, K.}, \bibinfo{author}{Reynolds, M.}, \bibinfo{author}{Ring, R.}, \bibinfo{author}{Rutherford, E.}, \bibinfo{author}{Cabi, S.}, \bibinfo{author}{Han, T.}, \bibinfo{author}{Gong, Z.}, \bibinfo{author}{Samangooei, S.}, \bibinfo{author}{Monteiro, M.}, \bibinfo{author}{Menick, J.}, \bibinfo{author}{Borgeaud, S.}, \bibinfo{author}{Brock, A.}, \bibinfo{author}{Nematzadeh, A.}, \bibinfo{author}{Sharifzadeh, S.}, \bibinfo{author}{Binkowski, M.}, \bibinfo{author}{Barreira, R.}, \bibinfo{author}{Vinyals, O.}, \bibinfo{author}{Zisserman, A.}, \bibinfo{author}{Simonyan, K.}, \bibinfo{year}{2022}.
\newblock \bibinfo{title}{Flamingo: a visual language model for few-shot learning}.
\newblock \bibinfo{journal}{ArXiv} \bibinfo{volume}{abs/2204.14198}.
\bibitem[{Barz and Denzler(2019)}]{embeddings_jena}
\bibinfo{author}{Barz, B.}, \bibinfo{author}{Denzler, J.}, \bibinfo{year}{2019}.
\newblock \bibinfo{title}{Hierarchy-based image embeddings for semantic image retrieval}, in: \bibinfo{booktitle}{2019 IEEE Winter Conference on Applications of Computer Vision (WACV)}, \bibinfo{publisher}{IEEE}.
\newblock \URLprefix \url{http://dx.doi.org/10.1109/WACV.2019.00073}, \DOIprefix\doi{10.1109/wacv.2019.00073}.
\bibitem[{Bertinetto et~al.(2020)Bertinetto, Mueller, Tertikas, Samangooei and Lord}]{cls_hierarchies}
\bibinfo{author}{Bertinetto, L.}, \bibinfo{author}{Mueller, R.}, \bibinfo{author}{Tertikas, K.}, \bibinfo{author}{Samangooei, S.}, \bibinfo{author}{Lord, N.A.}, \bibinfo{year}{2020}.
\newblock \bibinfo{title}{Making better mistakes: Leveraging class hierarchies with deep networks}.
\newblock \URLprefix \url{https://arxiv.org/abs/1912.09393}, \href{http://arxiv.org/abs/1912.09393}{\tt arXiv:1912.09393}.
\bibitem[{Beucher et~al.(2022)Beucher, Rasmussen, Moeslund and Greve}]{acid_sulfate}
\bibinfo{author}{Beucher, A.}, \bibinfo{author}{Rasmussen, C.B.}, \bibinfo{author}{Moeslund, T.B.}, \bibinfo{author}{Greve, M.H.}, \bibinfo{year}{2022}.
\newblock \bibinfo{title}{Interpretation of convolutional neural networks for acid sulfate soil classification}.
\newblock \bibinfo{journal}{Frontiers in Environmental Science} \bibinfo{volume}{9}.
\newblock \URLprefix \url{https://www.frontiersin.org/journals/environmental-science/articles/10.3389/fenvs.2021.809995}, \DOIprefix\doi{10.3389/fenvs.2021.809995}.
\bibitem[{Brown et~al.(2020)Brown, Mann, Ryder, Subbiah, Kaplan, Dhariwal, Neelakantan, Shyam, Sastry, Askell, Agarwal, Herbert-Voss, Krueger, Henighan, Child, Ramesh, Ziegler, Wu, Winter, Hesse, Chen, Sigler, Litwin, Gray, Chess, Clark, Berner, McCandlish, Radford, Sutskever and Amodei}]{llm_fewshot}
\bibinfo{author}{Brown, T.}, \bibinfo{author}{Mann, B.}, \bibinfo{author}{Ryder, N.}, \bibinfo{author}{Subbiah, M.}, \bibinfo{author}{Kaplan, J.D.}, \bibinfo{author}{Dhariwal, P.}, \bibinfo{author}{Neelakantan, A.}, \bibinfo{author}{Shyam, P.}, \bibinfo{author}{Sastry, G.}, \bibinfo{author}{Askell, A.}, \bibinfo{author}{Agarwal, S.}, \bibinfo{author}{Herbert-Voss, A.}, \bibinfo{author}{Krueger, G.}, \bibinfo{author}{Henighan, T.}, \bibinfo{author}{Child, R.}, \bibinfo{author}{Ramesh, A.}, \bibinfo{author}{Ziegler, D.}, \bibinfo{author}{Wu, J.}, \bibinfo{author}{Winter, C.}, \bibinfo{author}{Hesse, C.}, \bibinfo{author}{Chen, M.}, \bibinfo{author}{Sigler, E.}, \bibinfo{author}{Litwin, M.}, \bibinfo{author}{Gray, S.}, \bibinfo{author}{Chess, B.}, \bibinfo{author}{Clark, J.}, \bibinfo{author}{Berner, C.}, \bibinfo{author}{McCandlish, S.}, \bibinfo{author}{Radford, A.}, \bibinfo{author}{Sutskever, I.}, \bibinfo{author}{Amodei, D.}, \bibinfo{year}{2020}.
\newblock \bibinfo{title}{Language models are few-shot learners}, in: \bibinfo{editor}{Larochelle, H.}, \bibinfo{editor}{Ranzato, M.}, \bibinfo{editor}{Hadsell, R.}, \bibinfo{editor}{Balcan, M.}, \bibinfo{editor}{Lin, H.} (Eds.), \bibinfo{booktitle}{Advances in Neural Information Processing Systems}, \bibinfo{publisher}{Curran Associates, Inc.}. pp. \bibinfo{pages}{1877--1901}.
\newblock \URLprefix \url{https://proceedings.neurips.cc/paper_files/paper/2020/file/1457c0d6bfcb4967418bfb8ac142f64a-Paper.pdf}.
\bibitem[{{Bundesanstalt für Geowissenschaften und Rohstoffe}()}]{soil_bgr}
\bibinfo{author}{{Bundesanstalt für Geowissenschaften und Rohstoffe}}, .
\newblock \bibinfo{title}{{Soils: Area and spatial information}}.
\newblock \bibinfo{howpublished}{\url{https://www.bgr.bund.de/EN/Themen/Boden/Flaechen-Rauminformationen/flaechen-rauminformationen_node.html}}.
\newblock \bibinfo{note}{Accessed: 2025-08-05}.
\bibitem[{Canhui et~al.(2023)Canhui, Yuteng, Cao, Honghong, Hengyue and Yinong}]{doc_layout}
\bibinfo{author}{Canhui, X.}, \bibinfo{author}{Yuteng, L.}, \bibinfo{author}{Cao, S.}, \bibinfo{author}{Honghong, Z.}, \bibinfo{author}{Hengyue, B.}, \bibinfo{author}{Yinong, C.}, \bibinfo{year}{2023}.
\newblock \bibinfo{title}{Him: hierarchical multimodal network for document layout analysis}.
\newblock \bibinfo{journal}{Applied Intelligence} \bibinfo{volume}{53}, \bibinfo{pages}{24314–24326}.
\newblock \URLprefix \url{https://doi.org/10.1007/s10489-023-04782-3}, \DOIprefix\doi{10.1007/s10489-023-04782-3}.
\bibitem[{Chen et~al.(2021)Chen, Fan and Panda}]{crossvit}
\bibinfo{author}{Chen, C.F.}, \bibinfo{author}{Fan, Q.}, \bibinfo{author}{Panda, R.}, \bibinfo{year}{2021}.
\newblock \bibinfo{title}{Crossvit: Cross-attention multi-scale vision transformer for image classification}.
\newblock \URLprefix \url{https://arxiv.org/abs/2103.14899}, \href{http://arxiv.org/abs/2103.14899}{\tt arXiv:2103.14899}.
\bibitem[{Deng et~al.(2010)Deng, Berg, Li and Fei-Fei}]{10k_categories}
\bibinfo{author}{Deng, J.}, \bibinfo{author}{Berg, A.C.}, \bibinfo{author}{Li, K.}, \bibinfo{author}{Fei-Fei, L.}, \bibinfo{year}{2010}.
\newblock \bibinfo{title}{What does classifying more than 10,000 image categories tell us?}, in: \bibinfo{editor}{Daniilidis, K.}, \bibinfo{editor}{Maragos, P.}, \bibinfo{editor}{Paragios, N.} (Eds.), \bibinfo{booktitle}{Computer Vision -- ECCV 2010}, \bibinfo{publisher}{Springer Berlin Heidelberg}, \bibinfo{address}{Berlin, Heidelberg}. pp. \bibinfo{pages}{71--84}.
\bibitem[{{European Court of Auditors}(2018)}]{desertification}
\bibinfo{author}{{European Court of Auditors}}, \bibinfo{year}{2018}.
\newblock \bibinfo{title}{Background paper: Desertification in the eu}.
\newblock \bibinfo{howpublished}{\url{https://www.eca.europa.eu/en/publications/BP_DESERTIFICATION}}.
\newblock \bibinfo{note}{Accessed: 1.02.2025}.
\bibitem[{Fernandez~Astudillo et~al.(2020)Fernandez~Astudillo, Ballesteros, Naseem, Blodgett and Florian}]{stack_transformer}
\bibinfo{author}{Fernandez~Astudillo, R.}, \bibinfo{author}{Ballesteros, M.}, \bibinfo{author}{Naseem, T.}, \bibinfo{author}{Blodgett, A.}, \bibinfo{author}{Florian, R.}, \bibinfo{year}{2020}.
\newblock \bibinfo{title}{Transition-based parsing with stack-transformers}, in: \bibinfo{editor}{Cohn, T.}, \bibinfo{editor}{He, Y.}, \bibinfo{editor}{Liu, Y.} (Eds.), \bibinfo{booktitle}{Findings of the Association for Computational Linguistics: EMNLP 2020}, \bibinfo{publisher}{Association for Computational Linguistics}, \bibinfo{address}{Online}. pp. \bibinfo{pages}{1001--1007}.
\newblock \URLprefix \url{https://aclanthology.org/2020.findings-emnlp.89/}, \DOIprefix\doi{10.18653/v1/2020.findings-emnlp.89}.
\bibitem[{Froger et~al.(2024)Froger, Tondini, Arrouays, Oorts, Poeplau, Wetterlind, Putku, Saby, Fantappiè, Styc, Chenu, Salomez, Callewaert, Vanwindekens, Huyghebaert, Herinckx, Heilek, {Sofie Harbo}, {De Carvalho Gomes}, Lázaro-López, {Antonio Rodriguez}, Pindral, Smreczak, Benő, Bakacsi, Teuling, {van Egmond}, Hutár, Pálka, Abrahám and Bispo}]{lucas}
\bibinfo{author}{Froger, C.}, \bibinfo{author}{Tondini, E.}, \bibinfo{author}{Arrouays, D.}, \bibinfo{author}{Oorts, K.}, \bibinfo{author}{Poeplau, C.}, \bibinfo{author}{Wetterlind, J.}, \bibinfo{author}{Putku, E.}, \bibinfo{author}{Saby, N.P.}, \bibinfo{author}{Fantappiè, M.}, \bibinfo{author}{Styc, Q.}, \bibinfo{author}{Chenu, C.}, \bibinfo{author}{Salomez, J.}, \bibinfo{author}{Callewaert, S.}, \bibinfo{author}{Vanwindekens, F.M.}, \bibinfo{author}{Huyghebaert, B.}, \bibinfo{author}{Herinckx, J.}, \bibinfo{author}{Heilek, S.}, \bibinfo{author}{{Sofie Harbo}, L.}, \bibinfo{author}{{De Carvalho Gomes}, L.}, \bibinfo{author}{Lázaro-López, A.}, \bibinfo{author}{{Antonio Rodriguez}, J.}, \bibinfo{author}{Pindral, S.}, \bibinfo{author}{Smreczak, B.}, \bibinfo{author}{Benő, A.}, \bibinfo{author}{Bakacsi, Z.}, \bibinfo{author}{Teuling, K.}, \bibinfo{author}{{van Egmond}, F.}, \bibinfo{author}{Hutár, V.}, \bibinfo{author}{Pálka, B.}, \bibinfo{author}{Abrahám, D.}, \bibinfo{author}{Bispo, A.},
  \bibinfo{year}{2024}.
\newblock \bibinfo{title}{Comparing lucas soil and national systems: Towards a harmonized european soil monitoring network}.
\newblock \bibinfo{journal}{Geoderma} \bibinfo{volume}{449}, \bibinfo{pages}{117027}.
\newblock \URLprefix \url{https://www.sciencedirect.com/science/article/pii/S0016706124002568}, \DOIprefix\doi{https://doi.org/10.1016/j.geoderma.2024.117027}.
\bibitem[{{Google}(2025a)}]{gemini2025}
\bibinfo{author}{{Google}}, \bibinfo{year}{2025}a.
\newblock \bibinfo{title}{Gemini 2.0 flash}.
\newblock \URLprefix \url{https://gemini.google.com/app/}. \bibinfo{note}{accessed between 2025-05-05 and 2025-05-11}.
\bibitem[{{Google}(2025b)}]{geminiChatHistory2025}
\bibinfo{author}{{Google}}, \bibinfo{year}{2025}b.
\newblock \bibinfo{title}{Gemini chat history (public share)}.
\newblock \URLprefix \url{https://g.co/gemini/share/0ad1d2bf5561}. \bibinfo{note}{publicly shared chat history, accessed between 2025-05-05 and 2025-05-11}.
\bibitem[{Gu et~al.(2024)Gu, Zhang, Wang, Chen, Wang, Ge, Jiao, Ye, Jia and Wang}]{med_img_metadata}
\bibinfo{author}{Gu, R.}, \bibinfo{author}{Zhang, Y.}, \bibinfo{author}{Wang, L.}, \bibinfo{author}{Chen, D.}, \bibinfo{author}{Wang, Y.}, \bibinfo{author}{Ge, R.}, \bibinfo{author}{Jiao, Z.}, \bibinfo{author}{Ye, J.}, \bibinfo{author}{Jia, G.}, \bibinfo{author}{Wang, L.}, \bibinfo{year}{2024}.
\newblock \bibinfo{title}{Mmy-net: a multimodal network exploiting image and patient metadata for simultaneous segmentation and diagnosis}.
\newblock \bibinfo{journal}{Multimedia Syst.} \bibinfo{volume}{30}.
\newblock \URLprefix \url{https://doi.org/10.1007/s00530-024-01260-9}, \DOIprefix\doi{10.1007/s00530-024-01260-9}.
\bibitem[{Gupta and Kembhavi(2022)}]{visprog}
\bibinfo{author}{Gupta, T.}, \bibinfo{author}{Kembhavi, A.}, \bibinfo{year}{2022}.
\newblock \bibinfo{title}{Visual programming: Compositional visual reasoning without training}.
\newblock \bibinfo{journal}{ArXiv} \bibinfo{volume}{abs/2211.11559}.
\bibitem[{Gyasi and Purushotham(2023)}]{soil_review1}
\bibinfo{author}{Gyasi, E.K.}, \bibinfo{author}{Purushotham, S.}, \bibinfo{year}{2023}.
\newblock \bibinfo{title}{Advancements in soil classification: An in-depth analysis of current deep learning techniques and emerging trends}.
\newblock \bibinfo{journal}{Air, Soil and Water Research} \bibinfo{volume}{16}, \bibinfo{pages}{11786221231214069}.
\newblock \URLprefix \url{https://doi.org/10.1177/11786221231214069}, \DOIprefix\doi{10.1177/11786221231214069}, \href{http://arxiv.org/abs/https://doi.org/10.1177/11786221231214069}{\tt arXiv:https://doi.org/10.1177/11786221231214069}.
\bibitem[{Hartemink et~al.(2020)Hartemink, Zhang, Bockheim, Curi, Silva, Grauer-Gray, Lowe and Krasilnikov}]{HARTEMINK2020125}
\bibinfo{author}{Hartemink, A.}, \bibinfo{author}{Zhang, Y.}, \bibinfo{author}{Bockheim, J.}, \bibinfo{author}{Curi, N.}, \bibinfo{author}{Silva, S.}, \bibinfo{author}{Grauer-Gray, J.}, \bibinfo{author}{Lowe, D.}, \bibinfo{author}{Krasilnikov, P.}, \bibinfo{year}{2020}.
\newblock \bibinfo{title}{Chapter three - soil horizon variation: A review}, \bibinfo{publisher}{Academic Press}. volume \bibinfo{volume}{160} of \textit{\bibinfo{series}{Advances in Agronomy}}, pp. \bibinfo{pages}{125--185}.
\newblock \URLprefix \url{https://www.sciencedirect.com/science/article/pii/S0065211319301087}, \DOIprefix\doi{https://doi.org/10.1016/bs.agron.2019.10.003}.
\bibitem[{He et~al.(2015)He, Zhang, Ren and Sun}]{resnet}
\bibinfo{author}{He, K.}, \bibinfo{author}{Zhang, X.}, \bibinfo{author}{Ren, S.}, \bibinfo{author}{Sun, J.}, \bibinfo{year}{2015}.
\newblock \bibinfo{title}{Deep residual learning for image recognition}.
\newblock \URLprefix \url{https://arxiv.org/abs/1512.03385}, \DOIprefix\doi{10.48550/ARXIV.1512.03385}.
\bibitem[{Hengl et~al.(2007)Hengl, Heuvelink and Rossiter}]{resid_kriging}
\bibinfo{author}{Hengl, T.}, \bibinfo{author}{Heuvelink, G.B.}, \bibinfo{author}{Rossiter, D.G.}, \bibinfo{year}{2007}.
\newblock \bibinfo{title}{About regression-kriging: From equations to case studies}.
\newblock \bibinfo{journal}{Computers and Geosciences} \bibinfo{volume}{33}, \bibinfo{pages}{1301--1315}.
\newblock \URLprefix \url{https://www.sciencedirect.com/science/article/pii/S0098300407001008}, \DOIprefix\doi{https://doi.org/10.1016/j.cageo.2007.05.001}. \bibinfo{note}{spatial Analysis}.
\bibitem[{Hiller et~al.(2024)Hiller, Ehinger and Drummond}]{bixt}
\bibinfo{author}{Hiller, M.}, \bibinfo{author}{Ehinger, K.A.}, \bibinfo{author}{Drummond, T.}, \bibinfo{year}{2024}.
\newblock \bibinfo{title}{Perceiving longer sequences with bi-directional cross-attention transformers}.
\newblock \URLprefix \url{https://arxiv.org/abs/2402.12138}, \href{http://arxiv.org/abs/2402.12138}{\tt arXiv:2402.12138}.
\bibitem[{Hinterwimmer et~al.(2024)Hinterwimmer, Guenther, Consalvo, Neumann, Gersing, Woertler, {von Eisenhart-Rothe}, Burgkart and Rueckert}]{bone_tumor}
\bibinfo{author}{Hinterwimmer, F.}, \bibinfo{author}{Guenther, M.}, \bibinfo{author}{Consalvo, S.}, \bibinfo{author}{Neumann, J.}, \bibinfo{author}{Gersing, A.}, \bibinfo{author}{Woertler, K.}, \bibinfo{author}{{von Eisenhart-Rothe}, R.}, \bibinfo{author}{Burgkart, R.}, \bibinfo{author}{Rueckert, D.}, \bibinfo{year}{2024}.
\newblock \bibinfo{title}{Impact of metadata in multimodal classification of bone tumours}.
\newblock \bibinfo{journal}{BMC Musculoskeletal Disorders} \bibinfo{volume}{25}.
\newblock \DOIprefix\doi{10.1186/s12891-024-07934-9}. \bibinfo{note}{publisher Copyright: {\textcopyright} The Author(s) 2024.}
\bibitem[{Huang et~al.(2023)Huang, Dong, Wang, Hao, Singhal, Ma, Lv, Cui, Mohammed, Patra, Liu, Aggarwal, Chi, Bjorck, Chaudhary, Som, Song and Wei}]{kosmos}
\bibinfo{author}{Huang, S.}, \bibinfo{author}{Dong, L.}, \bibinfo{author}{Wang, W.}, \bibinfo{author}{Hao, Y.}, \bibinfo{author}{Singhal, S.}, \bibinfo{author}{Ma, S.}, \bibinfo{author}{Lv, T.}, \bibinfo{author}{Cui, L.}, \bibinfo{author}{Mohammed, O.K.}, \bibinfo{author}{Patra, B.}, \bibinfo{author}{Liu, Q.}, \bibinfo{author}{Aggarwal, K.}, \bibinfo{author}{Chi, Z.}, \bibinfo{author}{Bjorck, J.}, \bibinfo{author}{Chaudhary, V.}, \bibinfo{author}{Som, S.}, \bibinfo{author}{Song, X.}, \bibinfo{author}{Wei, F.}, \bibinfo{year}{2023}.
\newblock \bibinfo{title}{Language is not all you need: Aligning perception with language models}.
\newblock \URLprefix \url{https://arxiv.org/abs/2302.14045}, \href{http://arxiv.org/abs/2302.14045}{\tt arXiv:2302.14045}.
\bibitem[{{IUSS Working Group WRB}(2022)}]{iussWorldReferenceSoilResources}
\bibinfo{author}{{IUSS Working Group WRB}}, \bibinfo{year}{2022}.
\newblock \bibinfo{title}{World Reference Base for Soil Resources: International soil classification system for naming soils and creating legends for soil maps}.
\newblock \bibinfo{edition}{4th} ed., \bibinfo{publisher}{International Union of Soil Sciences (IUSS)}, \bibinfo{address}{Vienna, Austria}.
\newblock \URLprefix \url{https://wrb.isric.org/files/WRB_fourth_edition_2022-12-18_errata_correction_2024-09-24.pdf}.
\bibitem[{Jiang et~al.(2021)Jiang, Owens, Zhang, Brye, Weindorf, Adhikari, Sun, Sun and Wang}]{soil_survey}
\bibinfo{author}{Jiang, Z.D.}, \bibinfo{author}{Owens, P.R.}, \bibinfo{author}{Zhang, C.L.}, \bibinfo{author}{Brye, K.R.}, \bibinfo{author}{Weindorf, D.C.}, \bibinfo{author}{Adhikari, K.}, \bibinfo{author}{Sun, Z.X.}, \bibinfo{author}{Sun, F.J.}, \bibinfo{author}{Wang, Q.B.}, \bibinfo{year}{2021}.
\newblock \bibinfo{title}{Towards a dynamic soil survey: Identifying and delineating soil horizons in-situ using deep learning}.
\newblock \bibinfo{journal}{Geoderma} \bibinfo{volume}{401}, \bibinfo{pages}{115341}.
\newblock \URLprefix \url{https://www.sciencedirect.com/science/article/pii/S0016706121004213}, \DOIprefix\doi{https://doi.org/10.1016/j.geoderma.2021.115341}.
\bibitem[{Jiao et~al.(2019)Jiao, Sun, Yang, Ren and Liu}]{sattelite_multi_hyper}
\bibinfo{author}{Jiao, L.}, \bibinfo{author}{Sun, W.}, \bibinfo{author}{Yang, G.}, \bibinfo{author}{Ren, G.}, \bibinfo{author}{Liu, Y.}, \bibinfo{year}{2019}.
\newblock \bibinfo{title}{A hierarchical classification framework of satellite multispectral/hyperspectral images for mapping coastal wetlands}.
\newblock \bibinfo{journal}{Remote Sensing} \bibinfo{volume}{11}.
\newblock \URLprefix \url{https://www.mdpi.com/2072-4292/11/19/2238}, \DOIprefix\doi{10.3390/rs11192238}.
\bibitem[{Kirillov et~al.(2023)Kirillov, Mintun, Ravi, Mao, Rolland, Gustafson, Xiao, Whitehead, Berg, Lo, Dollár and Girshick}]{sam}
\bibinfo{author}{Kirillov, A.}, \bibinfo{author}{Mintun, E.}, \bibinfo{author}{Ravi, N.}, \bibinfo{author}{Mao, H.}, \bibinfo{author}{Rolland, C.}, \bibinfo{author}{Gustafson, L.}, \bibinfo{author}{Xiao, T.}, \bibinfo{author}{Whitehead, S.}, \bibinfo{author}{Berg, A.C.}, \bibinfo{author}{Lo, W.Y.}, \bibinfo{author}{Dollár, P.}, \bibinfo{author}{Girshick, R.}, \bibinfo{year}{2023}.
\newblock \bibinfo{title}{Segment anything}.
\newblock \URLprefix \url{https://arxiv.org/abs/2304.02643}, \href{http://arxiv.org/abs/2304.02643}{\tt arXiv:2304.02643}.
\bibitem[{Kirillova et~al.(2015)Kirillova, Vodyanitskii and Sileva}]{cie_lab}
\bibinfo{author}{Kirillova, N.}, \bibinfo{author}{Vodyanitskii, Y.}, \bibinfo{author}{Sileva, T.}, \bibinfo{year}{2015}.
\newblock \bibinfo{title}{Conversion of soil color parameters from the munsell system to the cie-l*a*b* system}.
\newblock \bibinfo{journal}{Eurasian Soil Science} \bibinfo{volume}{48}, \bibinfo{pages}{527}.
\newblock \DOIprefix\doi{10.1134/S1064229315050026}.
\bibitem[{Kühn and Eberhardt(2023)}]{kühn_eberhardt_soil_texture}
\bibinfo{author}{Kühn, D.}, \bibinfo{author}{Eberhardt, E.}, \bibinfo{year}{2023}.
\newblock \bibinfo{title}{A classification of soil solid material for natural and anthropogenic soils}.
\newblock \bibinfo{journal}{Journal of Plant Nutrition and Soil Science} \bibinfo{volume}{186}, \bibinfo{pages}{507--521}.
\newblock \URLprefix \url{https://onlinelibrary.wiley.com/doi/abs/10.1002/jpln.202200444}, \DOIprefix\doi{https://doi.org/10.1002/jpln.202200444}, \href{http://arxiv.org/abs/https://onlinelibrary.wiley.com/doi/pdf/10.1002/jpln.202200444}{\tt arXiv:https://onlinelibrary.wiley.com/doi/pdf/10.1002/jpln.202200444}.
\bibitem[{Lagacherie(2008)}]{lagacherieDigitalSoilMappingSOTA}
\bibinfo{author}{Lagacherie, P.}, \bibinfo{year}{2008}.
\newblock \bibinfo{title}{Digital Soil Mapping: A State of the Art}. \bibinfo{publisher}{Springer Netherlands}, \bibinfo{address}{Dordrecht}.
\newblock pp. \bibinfo{pages}{3--14}.
\newblock \URLprefix \url{https://doi.org/10.1007/978-1-4020-8592-5\_1}, \DOIprefix\doi{10.1007/978-1-4020-8592-5\_1}.
\bibitem[{Li et~al.(2022)Li, Zhou, Wang, Li and Yang}]{liDeepHierarchicalSemantic2022}
\bibinfo{author}{Li, L.}, \bibinfo{author}{Zhou, T.}, \bibinfo{author}{Wang, W.}, \bibinfo{author}{Li, J.}, \bibinfo{author}{Yang, Y.}, \bibinfo{year}{2022}.
\newblock \bibinfo{title}{Deep {Hierarchical} {Semantic} {Segmentation}}, pp. \bibinfo{pages}{1246--1257}.
\newblock \URLprefix \url{https://openaccess.thecvf.com/content/CVPR2022/html/Li_Deep_Hierarchical_Semantic_Segmentation_CVPR_2022_paper.html}.
\bibitem[{Lin et~al.(2022)Lin, Cheng, Wu and Shen}]{cat}
\bibinfo{author}{Lin, H.}, \bibinfo{author}{Cheng, X.}, \bibinfo{author}{Wu, X.}, \bibinfo{author}{Shen, D.}, \bibinfo{year}{2022}.
\newblock \bibinfo{title}{Cat: Cross attention in vision transformer}, in: \bibinfo{booktitle}{2022 IEEE International Conference on Multimedia and Expo (ICME)}, pp. \bibinfo{pages}{1--6}.
\newblock \DOIprefix\doi{10.1109/ICME52920.2022.9859720}.
\bibitem[{Lin et~al.(2024)Lin, Li, Chen, Xu, Clark and Torr}]{olympus}
\bibinfo{author}{Lin, Y.}, \bibinfo{author}{Li, Y.}, \bibinfo{author}{Chen, D.}, \bibinfo{author}{Xu, W.}, \bibinfo{author}{Clark, R.}, \bibinfo{author}{Torr, P.H.}, \bibinfo{year}{2024}.
\newblock \bibinfo{title}{Olympus: A universal task router for computer vision tasks}.
\newblock \bibinfo{journal}{arXiv preprint arXiv:2412.09612} .
\bibitem[{McBratney et~al.(2003)McBratney, {Mendonça Santos} and Minasny}]{mcbratneyDigitalSoilMapping}
\bibinfo{author}{McBratney, A.}, \bibinfo{author}{{Mendonça Santos}, M.}, \bibinfo{author}{Minasny, B.}, \bibinfo{year}{2003}.
\newblock \bibinfo{title}{On digital soil mapping}.
\newblock \bibinfo{journal}{Geoderma} \bibinfo{volume}{117}, \bibinfo{pages}{3--52}.
\newblock \URLprefix \url{https://www.sciencedirect.com/science/article/pii/S0016706103002234}, \DOIprefix\doi{10.1016/S0016-7061(03)00223-4}.
\bibitem[{Mihaylova and Martins(2019)}]{teacherforcing1}
\bibinfo{author}{Mihaylova, T.}, \bibinfo{author}{Martins, A.F.T.}, \bibinfo{year}{2019}.
\newblock \bibinfo{title}{Scheduled sampling for transformers}, in: \bibinfo{editor}{Alva-Manchego, F.}, \bibinfo{editor}{Choi, E.}, \bibinfo{editor}{Khashabi, D.} (Eds.), \bibinfo{booktitle}{Proceedings of the 57th Annual Meeting of the Association for Computational Linguistics: Student Research Workshop}, \bibinfo{publisher}{Association for Computational Linguistics}, \bibinfo{address}{Florence, Italy}. pp. \bibinfo{pages}{351--356}.
\newblock \URLprefix \url{https://aclanthology.org/P19-2049/}, \DOIprefix\doi{10.18653/v1/P19-2049}.
\bibitem[{Nickel and Kiela(2017)}]{nickelPoincareEmbeddingsLearning2017}
\bibinfo{author}{Nickel, M.}, \bibinfo{author}{Kiela, D.}, \bibinfo{year}{2017}.
\newblock \bibinfo{title}{Poincaré {Embeddings} for {Learning} {Hierarchical} {Representations}}.
\newblock \URLprefix \url{http://arxiv.org/abs/1705.08039}, \DOIprefix\doi{10.48550/arXiv.1705.08039}. \bibinfo{note}{arXiv:1705.08039 [cs]}.
\bibitem[{de~Oliveira et~al.(2023)de~Oliveira, Falcioni, Gonçalves, de~Oliveira, Mendonça, Crusiol, de~Oliveira, Furlanetto, Reis and Nanni}]{soil_sprectroscopy}
\bibinfo{author}{de~Oliveira, K.M.}, \bibinfo{author}{Falcioni, R.}, \bibinfo{author}{Gonçalves, J.V.F.}, \bibinfo{author}{de~Oliveira, C.A.}, \bibinfo{author}{Mendonça, W.A.}, \bibinfo{author}{Crusiol, L.G.T.}, \bibinfo{author}{de~Oliveira, R.B.}, \bibinfo{author}{Furlanetto, R.H.}, \bibinfo{author}{Reis, A.S.}, \bibinfo{author}{Nanni, M.R.}, \bibinfo{year}{2023}.
\newblock \bibinfo{title}{Rapid determination of soil horizons and suborders based on vis-nir-swir spectroscopy and machine learning models}.
\newblock \bibinfo{journal}{Remote Sensing} \bibinfo{volume}{15}.
\newblock \URLprefix \url{https://www.mdpi.com/2072-4292/15/19/4859}, \DOIprefix\doi{10.3390/rs15194859}.
\bibitem[{{OpenAI}(2025a)}]{chatgpt4omini2025}
\bibinfo{author}{{OpenAI}}, \bibinfo{year}{2025}a.
\newblock \bibinfo{title}{Chatgpt-4o mini}.
\newblock \URLprefix \url{https://chat.openai.com/chat}. \bibinfo{note}{accessed between 2025-05-05 and 2025-05-11}.
\bibitem[{{OpenAI}(2025b)}]{chatgptChatHistory2025}
\bibinfo{author}{{OpenAI}}, \bibinfo{year}{2025}b.
\newblock \bibinfo{title}{Chatgpt chat history (public share)}.
\newblock \URLprefix \url{https://chatgpt.com/share/681f47c2-8484-800d-bdc4-ec5c10e72b4e}. \bibinfo{note}{publicly shared chat history, accessed between 2025-05-05 and 2025-05-11}.
\bibitem[{Prabhavathi and Kuppusamy(2022)}]{soil_review2}
\bibinfo{author}{Prabhavathi, V.}, \bibinfo{author}{Kuppusamy, P.}, \bibinfo{year}{2022}.
\newblock \bibinfo{title}{A study on deep learning based soil classification}, in: \bibinfo{booktitle}{2022 IEEE 4th International Conference on Cybernetics, Cognition and Machine Learning Applications (ICCCMLA)}, pp. \bibinfo{pages}{428--433}.
\newblock \DOIprefix\doi{10.1109/ICCCMLA56841.2022.9989293}.
\bibitem[{Ronaldo(2021)}]{soil_cnn}
\bibinfo{author}{Ronaldo, A.}, \bibinfo{year}{2021}.
\newblock \bibinfo{title}{Effective soil type classification using convolutional neural network}.
\newblock \bibinfo{journal}{International Journal of Informatics and Computation} \bibinfo{volume}{3}, \bibinfo{pages}{20}.
\newblock \DOIprefix\doi{10.35842/ijicom.v3i1.33}.
\bibitem[{Sak et~al.(2014)Sak, Senior and Beaufays}]{lstm}
\bibinfo{author}{Sak, H.}, \bibinfo{author}{Senior, A.}, \bibinfo{author}{Beaufays, F.}, \bibinfo{year}{2014}.
\newblock \bibinfo{title}{Long short-term memory based recurrent neural network architectures for large vocabulary speech recognition}.
\newblock \URLprefix \url{https://arxiv.org/abs/1402.1128}, \href{http://arxiv.org/abs/1402.1128}{\tt arXiv:1402.1128}.
\bibitem[{Sechidis et~al.(2011)Sechidis, Tsoumakas and Vlahavas}]{multistratified}
\bibinfo{author}{Sechidis, K.}, \bibinfo{author}{Tsoumakas, G.}, \bibinfo{author}{Vlahavas, I.}, \bibinfo{year}{2011}.
\newblock \bibinfo{title}{On the stratification of multi-label data}, in: \bibinfo{editor}{Gunopulos, D.}, \bibinfo{editor}{Hofmann, T.}, \bibinfo{editor}{Malerba, D.}, \bibinfo{editor}{Vazirgiannis, M.} (Eds.), \bibinfo{booktitle}{Machine Learning and Knowledge Discovery in Databases}, \bibinfo{publisher}{Springer Berlin Heidelberg}, \bibinfo{address}{Berlin, Heidelberg}. pp. \bibinfo{pages}{145--158}.
\bibitem[{Shah et~al.(2018)Shah, Zhou, Abrámoff and Wu}]{seg_retina}
\bibinfo{author}{Shah, A.}, \bibinfo{author}{Zhou, L.}, \bibinfo{author}{Abrámoff, M.}, \bibinfo{author}{Wu, X.}, \bibinfo{year}{2018}.
\newblock \bibinfo{title}{Multiple surface segmentation using convolution neural nets: application to retinal layer segmentation in oct images}.
\newblock \bibinfo{journal}{Biomedical Optics Express} \bibinfo{volume}{9}, \bibinfo{pages}{4509--4526}.
\newblock \DOIprefix\doi{10.1364/BOE.9.004509}.
\bibitem[{Shen et~al.(2023)Shen, Song, Tan, Li, Lu and Zhuang}]{hugginggpt}
\bibinfo{author}{Shen, Y.}, \bibinfo{author}{Song, K.}, \bibinfo{author}{Tan, X.}, \bibinfo{author}{Li, D.}, \bibinfo{author}{Lu, W.}, \bibinfo{author}{Zhuang, Y.}, \bibinfo{year}{2023}.
\newblock \bibinfo{title}{Hugginggpt: Solving ai tasks with chatgpt and its friends in huggingface}, in: \bibinfo{booktitle}{Advances in Neural Information Processing Systems}.
\bibitem[{Spaargaren and Deckers(1998)}]{world_soils}
\bibinfo{author}{Spaargaren, O.C.}, \bibinfo{author}{Deckers, J.}, \bibinfo{year}{1998}.
\newblock \bibinfo{title}{The world reference base for soil resources}, in: \bibinfo{editor}{Schulte, A.}, \bibinfo{editor}{Ruhiyat, D.} (Eds.), \bibinfo{booktitle}{Soils of Tropical Forest Ecosystems}, \bibinfo{publisher}{Springer Berlin Heidelberg}, \bibinfo{address}{Berlin, Heidelberg}. pp. \bibinfo{pages}{21--28}.
\bibitem[{Surís et~al.(2023)Surís, Menon and Vondrick}]{vipergpt}
\bibinfo{author}{Surís, D.}, \bibinfo{author}{Menon, S.}, \bibinfo{author}{Vondrick, C.}, \bibinfo{year}{2023}.
\newblock \bibinfo{title}{Vipergpt: Visual inference via python execution for reasoning}.
\newblock \URLprefix \url{https://arxiv.org/abs/2303.08128}, \href{http://arxiv.org/abs/2303.08128}{\tt arXiv:2303.08128}.
\bibitem[{{Thünen Institute}()}]{soil_fotos}
\bibinfo{author}{{Thünen Institute}}, .
\newblock \bibinfo{title}{{German Agricultural Soil Inventory (BZE-LW)}}.
\newblock \bibinfo{howpublished}{\url{https://www.thuenen.de/en/institutes/climate-smart-agriculture/projects/agricultural-soil-inventory-bze-lw}}.
\newblock \bibinfo{note}{Accessed: 2025-08-05}.
\bibitem[{Tresson et~al.(2024)Tresson, Dumont, Jaeger, Borne, Boivin, Marie-Louise, François, Boukcim and Goëau}]{soil_selfsupervised_vit}
\bibinfo{author}{Tresson, P.}, \bibinfo{author}{Dumont, M.}, \bibinfo{author}{Jaeger, M.}, \bibinfo{author}{Borne, F.}, \bibinfo{author}{Boivin, S.}, \bibinfo{author}{Marie-Louise, L.}, \bibinfo{author}{François, J.}, \bibinfo{author}{Boukcim, H.}, \bibinfo{author}{Goëau, H.}, \bibinfo{year}{2024}.
\newblock \bibinfo{title}{Self-supervised learning of vision transformers for digital soil mapping using visual data}.
\newblock \bibinfo{journal}{Geoderma} \bibinfo{volume}{450}, \bibinfo{pages}{117056}.
\newblock \URLprefix \url{https://www.sciencedirect.com/science/article/pii/S0016706124002854}, \DOIprefix\doi{https://doi.org/10.1016/j.geoderma.2024.117056}.
\bibitem[{Turk(2023)}]{Turk_jellyfish_2023}
\bibinfo{author}{Turk, J.}, \bibinfo{year}{2023}.
\newblock \bibinfo{title}{{jellyfish}}.
\newblock \URLprefix \url{https://github.com/jamesturk/jellyfish}.
\bibitem[{Wadoux et~al.(2020)Wadoux, Minasny and McBratney}]{wadouxMachineLearningDigitalSoilMapping}
\bibinfo{author}{Wadoux, A.M.C.}, \bibinfo{author}{Minasny, B.}, \bibinfo{author}{McBratney, A.B.}, \bibinfo{year}{2020}.
\newblock \bibinfo{title}{Machine learning for digital soil mapping: Applications, challenges and suggested solutions}.
\newblock \bibinfo{journal}{Earth-Science Reviews} \bibinfo{volume}{210}, \bibinfo{pages}{103359}.
\newblock \URLprefix \url{https://www.sciencedirect.com/science/article/pii/S0012825220304050}, \DOIprefix\doi{10.1016/j.earscirev.2020.103359}.
\bibitem[{Yamagata et~al.(2024)Yamagata, Ikeno, Kimura, Isokawa, Nakaji, Mori, Kume and Ohashi}]{root_segmentation}
\bibinfo{author}{Yamagata, T.}, \bibinfo{author}{Ikeno, H.}, \bibinfo{author}{Kimura, T.}, \bibinfo{author}{Isokawa, T.}, \bibinfo{author}{Nakaji, T.}, \bibinfo{author}{Mori, K.}, \bibinfo{author}{Kume, T.}, \bibinfo{author}{Ohashi, M.}, \bibinfo{year}{2024}.
\newblock \bibinfo{title}{Segmenting growing and dying woody roots in a forest using deep learning-based misc}, in: \bibinfo{booktitle}{TENCON 2024 - 2024 IEEE Region 10 Conference (TENCON)}, pp. \bibinfo{pages}{535--538}.
\newblock \DOIprefix\doi{10.1109/TENCON61640.2024.10902758}.
\bibitem[{Yan et~al.(2015)Yan, Zhang, Piramuthu, Jagadeesh, DeCoste, Di and Yu}]{hdcnn}
\bibinfo{author}{Yan, Z.}, \bibinfo{author}{Zhang, H.}, \bibinfo{author}{Piramuthu, R.}, \bibinfo{author}{Jagadeesh, V.}, \bibinfo{author}{DeCoste, D.}, \bibinfo{author}{Di, W.}, \bibinfo{author}{Yu, Y.}, \bibinfo{year}{2015}.
\newblock \bibinfo{title}{Hd-cnn: Hierarchical deep convolutional neural network for large scale visual recognition}.
\newblock \URLprefix \url{https://arxiv.org/abs/1410.0736}, \href{http://arxiv.org/abs/1410.0736}{\tt arXiv:1410.0736}.
\bibitem[{Zhang and Hartemink(2019)}]{soil_horizon_kmeans}
\bibinfo{author}{Zhang, Y.}, \bibinfo{author}{Hartemink, A.E.}, \bibinfo{year}{2019}.
\newblock \bibinfo{title}{A method for automated soil horizon delineation using digital images}.
\newblock \bibinfo{journal}{Geoderma} \bibinfo{volume}{343}, \bibinfo{pages}{97--115}.
\newblock \URLprefix \url{https://www.sciencedirect.com/science/article/pii/S0016706118321141}, \DOIprefix\doi{https://doi.org/10.1016/j.geoderma.2019.02.002}.
\bibitem[{Zhou et~al.(2024)Zhou, Biswas, Hong, Chen, Hu, Shi, Guo and Li}]{soil_sprectroscopy_2}
\bibinfo{author}{Zhou, Y.}, \bibinfo{author}{Biswas, A.}, \bibinfo{author}{Hong, Y.}, \bibinfo{author}{Chen, S.}, \bibinfo{author}{Hu, B.}, \bibinfo{author}{Shi, Z.}, \bibinfo{author}{Guo, Y.}, \bibinfo{author}{Li, S.}, \bibinfo{year}{2024}.
\newblock \bibinfo{title}{Enhancing soil profile analysis with soil spectral libraries and laboratory hyperspectral imaging}.
\newblock \bibinfo{journal}{Geoderma} \bibinfo{volume}{450}, \bibinfo{pages}{117036}.
\newblock \URLprefix \url{https://www.sciencedirect.com/science/article/pii/S0016706124002659}, \DOIprefix\doi{https://doi.org/10.1016/j.geoderma.2024.117036}.
\bibitem[{Zhu and Bain(2017)}]{bcnn}
\bibinfo{author}{Zhu, X.}, \bibinfo{author}{Bain, M.}, \bibinfo{year}{2017}.
\newblock \bibinfo{title}{B-cnn: Branch convolutional neural network for hierarchical classification}.
\newblock \URLprefix \url{https://arxiv.org/abs/1709.09890}, \href{http://arxiv.org/abs/1709.09890}{\tt arXiv:1709.09890}.

\end{thebibliography}

\newpage
\appendix
\renewcommand{\thesection}{\Alph{section}} 

\begin{revisionsection}
\section{Corresponding Horizon Labels in WRB system}
In \autoref{tab:horizon_labels_ka5_wrb}, we provide the mapping between the horizon labels of the German KA5 system~\citep{kartieranleitung5} and those of the IUSS Working Group WRB system~\citep{iussWorldReferenceSoilResources}.
\begin{table}[pos=h]
    \caption{Mapping of horizon labels from the German KA5 classification system~\citep{kartieranleitung5} to their corresponding labels in the WRB system~\citep{iussWorldReferenceSoilResources}.}
    \label{tab:horizon_labels_ka5_wrb}
    \begin{revisionsection}
    \centering

    \begin{minipage}{0.32\textwidth}
        \centering
        \begin{tabular}{ll}
            \toprule
            \textbf{KA5} & \textbf{WRB} \\
            \midrule
            Ah & Ah \\
            Acxh & Ahk \\
            Axh & Ah \\
            Sw-Ah & Ahg \\
            Aeh & AhE, Eh \\
            Ael & E \\
            Bv-Ael & EBw \\
            Al & E \\
            Bt-Al & E/Bt \\
            Bv-Al & EBw \\
            Sw-Al & EBg \\
            Acp & Akp \\
            Aep & Ep \\
            Ap & Ap \\
            Axp & Ahp \\
            Sw-Ap & Apg \\
            Bv & Bw \\
            Bcv & Bwk \\
            Btv & Bwt \\
            Bhv & Bwh \\
            Bsv & Bws \\
            Ael-Bv & BwE \\
            Ah-Bv & BwA \\
            Al-Bv & BwE \\
            Cv-Bv & Bw \\
            ilC-Bv & BwC \\
            Sd-Btv & Btwg \\
            Sd-Bv & Bwg \\
            Ael-Bt & BtE \\
            Bvt & Btw \\
            Sd-Bt & Btg \\
            Sw-Bt & Btg \\
            \bottomrule
        \end{tabular}
    \end{minipage}
    \hfill
    \begin{minipage}{0.32\textwidth}
        \centering
        \begin{tabular}{ll}
            \toprule
            \textbf{KA5} & \textbf{WRB} \\
            \midrule
            Bsh & Bhs \\
            Bhs & Bsh \\
            Bs & Bs \\
            C & C, R \\
            iC & C \\
            ilC & C \\
            aelC & C\textalpha \\
            cC & C\textalpha, R\textalpha \\
            clC & C\textalpha \\
            elCc & Ck\textalpha \\
            elC & C\textalpha \\
            imC & R \\
            cmC & R\textalpha \\
            emC & R\textalpha \\
            Bv-ilC & CB \\
            Go-ilC & CBl \\
            Sw-ilC & Cg \\
            Sd-ilC & CBg \\
            Sd-lC & CBl \\
            Cj & C, Ce \\
            Bv-Cv & CB \\
            Bv-elCv & CaB \\
            Cbtv & Ct \\
            Cv & C \\
            ilCbtv & Ct \\
            imCv & R \\
            mCv & R \\
            bE & B \\
            E & B \\
            Go & Bl \\
            Gro & Bl \\
            Gkso & Blc \\
            \bottomrule
        \end{tabular}
    \end{minipage}
    \hfill
    \begin{minipage}{0.32\textwidth}
        \centering
        \begin{tabular}{ll}
            \toprule
            \textbf{KA5} & \textbf{WRB} \\
            \midrule
            M-Go & Bl \\
            M-Gro & Bl \\
            Sw-Go & Blg \\
            tGo & Bl \\
            tGro & Brl \\
            Gor & Blr \\
            Gr & Br \\
            tGr & Br \\
            Gw & B\textsigma \\
            Hr & H \\
            Hv & He \\
            Hw & H, Hi \\
            Go-M & Bl \\
            M & B \\
            Sw-M & Bg \\
            Mc & Bk \\
            elC-P & BwC\textalpha \\
            P & B, (Bi) \\
            R & B \\
            Sw & Bg \\
            Ssw & Bg \\
            Al-Sw & BgE \\
            Bv-Sw & Bgw \\
            M-Sw & Bg \\
            Sd & Bg \\
            Sswd & Bg \\
            Bt-Sd & Bgt \\
            Bv-Sd & Bgw \\
            ilC-Sd & BgC \\
            P-Sd & Bgw \\
            \\
            \\
            \bottomrule
        \end{tabular}
    \end{minipage}
    \end{revisionsection}

\end{table}

\section{User Study Screenshots}\label{sec:user_study_screenshots}
This section provides screenshots of the web-based annotation interface developed for our user study (\autoref{subsec:user_study_methods}). The frontend guides experts through drawing horizon boundaries by double-clicking on the image and assigning horizon symbols while presenting the hierarchical taxonomy used throughout the paper. \autoref{fig:user_study_screenshots} highlights the tutorial entry point, the hierarchical label selection, and the zoom/overview views that supports precise annotation.
The application texts are written in German to align with the language used by the participating soil surveyors.

\begin{figure*}[pos=h]
    \centering
    \subfloat[Welcome screen with tutorial]{\includegraphics[width=0.495\textwidth]{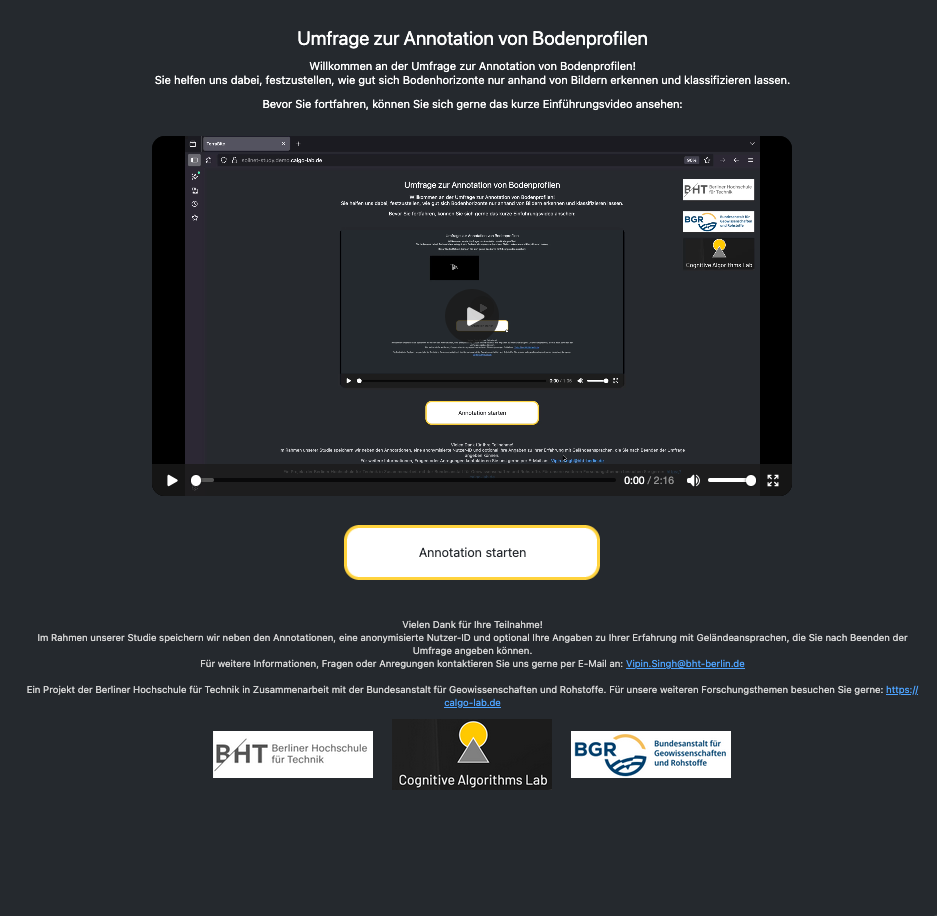}}
    \hfill    
    \subfloat[Full screen and zooming option]{\includegraphics[width=0.495\textwidth]{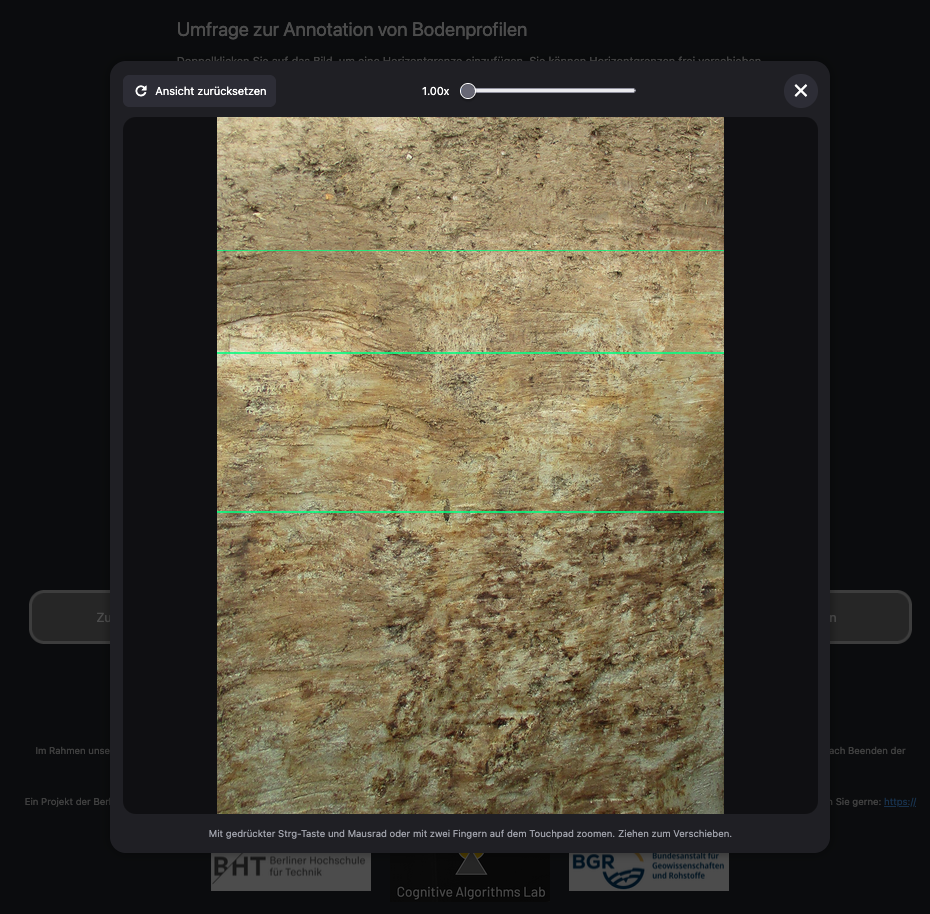}}
    
    \subfloat[Hierarchical selection of horizon symbols]{\includegraphics[width=0.495\textwidth]{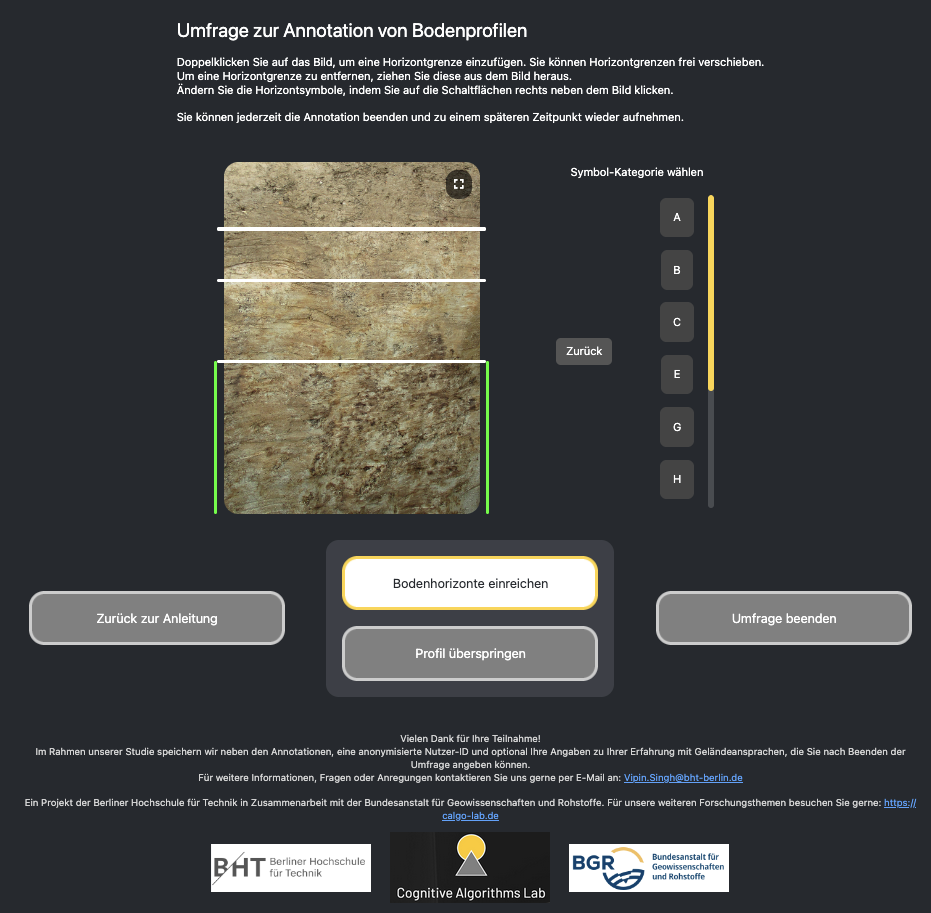}}
    \hfill
    \subfloat[Overview]{\includegraphics[width=0.495\textwidth]{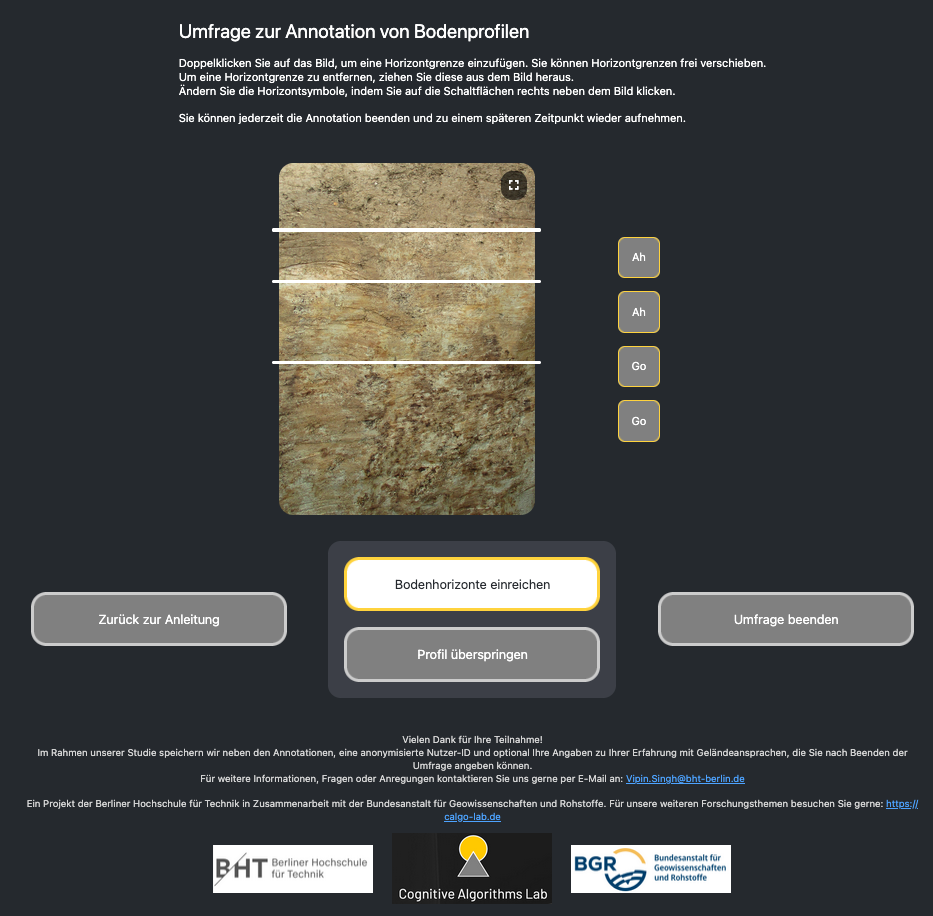}}
    \caption{\revisionchange{\textbf{Screenshots of the web frontend used in our user study.} The interface supports tutorial-based onboarding (a), zooming for precise boundary placement (b), hierarchical selection of horizon symbols consistent with our taxonomy (c), and an overview panel for reviewing the current segmentation and labels (d).}}
    \label{fig:user_study_screenshots}
\end{figure*}

\end{revisionsection}
\FloatBarrier
\section{Computing the Horizon Embeddings}\label{sec:ex_emb}

This section presents and proves the similarity guarantees offered by the horizon embeddings and walks the reader through a simple numerical example with 10 horizon embeddings.

\subsection{Similarity Invariances}
Depending on the order in which they are computed, the label embeddings may land in different subspaces within the embedding space $\mathbb{R}^{d_E}$. However, the algorithm guarantees invariant inner products between all categories of label combinations. Without loss of generality, let the equation of a (normalized) mixture embedding be written as:
\begin{equation}\label{eq:mixture_simplify}
    \varphi(h_3) = \frac{\frac{1}{3}\varphi(h_1) + \frac{2}{3}\varphi(h_2)}{||\frac{1}{3}\varphi(h_1) + \frac{2}{3}\varphi(h_2)||} = \frac{1}{\sqrt{5}}(\varphi(h_1) + 2\varphi(h_2))
\end{equation}
since $||\varphi(h_1)|| = ||\varphi(h_2)|| = 1$ and $\varphi(h_1) \perp \varphi(h_2)$. The embeddings of the mixture parents will always be orthogonal on each other because a mixture is always composed of labels containing different main symbols.

Then, the following inner product identities hold\footnote{For conciseness and readability, we will henceforth denote the dot product (or similarity) of the embeddings of two horizon labels simply as the dot product of those labels, while asking the reader to keep in mind that labels are strings and embeddings are vectors.}:

\begin{itemize}
    \item[\textbf{1)}] By definition, the dot product of any two labels (mixtures or non-mixtures) that do not have any main symbol in common is 0. Example: \textbf{Ah} vs \textbf{Bt}, \textbf{Ah-Bv} vs \textbf{Sd} or \textbf{Ah-Bv} vs \textbf{M-Sw}.
    
    \item[\textbf{2)}] Two non-mixture labels belonging to the same main symbol will always have similarity $\frac{1}{2}$ - see \autoref{eq:lca} and \autoref{eq:sim}. Example: \textbf{Al} vs \textbf{Ah.}
    
    \item[\textbf{3)}] Mixture vs Non-mixture:
    \begin{itemize}
        \item[3.1)] The non-mixture is one of the parents in the mixture: the dot product is $\frac{1}{\sqrt{5}}$, if the non-mixture label is identical to the first parent in the mixture e.g. \textbf{Al} vs \textbf{Al-Bt}, and $\frac{2}{\sqrt{5}}$, if the non-mixture is identical to the second (dominant) parent in the mixture e.g. \textbf{Bt} vs \textbf{Al-Bt}. 
        \begin{proof}
            Let $\varphi(h_3)$ be defined as in \autoref{eq:mixture_simplify}. Then:
            \begin{align*}
                \langle\varphi(h_1), \varphi(h_3)\rangle &= \tfrac{1}{\sqrt{5}}\langle\varphi(h_1), \varphi(h_1) + 2 \cdot \varphi(h_2)\rangle \\
                                                         &= \tfrac{1}{\sqrt{5}}(\underbrace{\langle\varphi(h_1), \varphi(h_1)\rangle}_\text{1} + 2 \cdot \underbrace{\langle\varphi(h_1), \varphi(h_2)\rangle}_\text{0})
                                                         = \tfrac{1}{\sqrt{5}}
            \end{align*}
            Analogously:
            \begin{align*}
                \langle\varphi(h_2), \varphi(h_3)\rangle &= \tfrac{1}{\sqrt{5}}\langle\varphi(h_2), \varphi(h_1) + 2 \cdot \varphi(h_2)\rangle \\
                                                         &= \tfrac{1}{\sqrt{5}}(\underbrace{\langle\varphi(h_2), \varphi(h_1)\rangle}_\text{0} + 2 \cdot \underbrace{\langle\varphi(h_2), \varphi(h_2)\rangle}_\text{1})
                                                         = \tfrac{2}{\sqrt{5}}
            \end{align*}
        \end{proof}
        \item[3.2)] The non-mixture is not one of the parents in the mixture, but has a common main symbol with one of the parents: the dot product is $\frac{1}{2\sqrt{5}}$, if the common main symbol comes from the first parent e.g. \textbf{Ah} vs \textbf{Al-Bt}, and $\frac{1}{\sqrt{5}}$ otherwise, e.g. \textbf{Bv} vs \textbf{Al-Bt}.
        \begin{proof}
            Let $\varphi(h_3)$ be defined as in \autoref{eq:mixture_simplify}. Then:
            \begin{align*}
                \langle\varphi(h_1), \varphi(h_3)\rangle &= \tfrac{1}{\sqrt{5}}\langle\varphi(h_1), \varphi(h_1) + 2 \cdot \varphi(h_2)\rangle \\
                                                         &= \tfrac{1}{\sqrt{5}}(\underbrace{\langle\varphi(h_1), \varphi(h_1)\rangle}_\text{1/2} + 2 \cdot \underbrace{\langle\varphi(h_1), \varphi(h_2)\rangle}_\text{0})
                                                         = \tfrac{1}{2\sqrt{5}}
            \end{align*}
            Analogously:
            \begin{align*}
                \langle\varphi(h_2), \varphi(h_3)\rangle &= \tfrac{1}{\sqrt{5}}\langle\varphi(h_2), \varphi(h_1) + 2 \cdot \varphi(h_2)\rangle \\
                                                         &= \tfrac{1}{\sqrt{5}}(\underbrace{\langle\varphi(h_2), \varphi(h_1)\rangle}_\text{0} + 2 \cdot \underbrace{\langle\varphi(h_2), \varphi(h_2)\rangle}_\text{1/2})
                                                         = \tfrac{1}{\sqrt{5}}
            \end{align*}
        \end{proof}
    \end{itemize}

    \item[\textbf{4)}] Mixture vs Mixture:
    \begin{itemize}
        \item[4.1)] The mixtures have one parent in common. Through analogous steps as the ones above, one arrives at the following results:
        \begin{itemize}
            \item[4.1.1)] If the common parent is at the second position e.g. \textbf{Ah-Bv} vs \textbf{Sd-Bv}, the dot product is $\frac{4}{5}$.
            \item[4.1.2)] If the common parent is at the first position e.g. \textbf{Bv-Cv} vs \textbf{Bv-Ael}, the dor product is $\frac{1}{5}$.
            \item[4.1.3)] If the common parent is at the first position in one mixture and second position in the other mixture e.g. \textbf{Ah-Bv} vs \textbf{Bv-Cv}, the dot product is $\frac{2}{5}$.
        \end{itemize}
        \item[4.2)] The mixtures have no common parent, but one common main symbol. Through analogous steps as the ones above, one arrives at the following results:
        \begin{itemize}
            \item[4.2.1)] One common main symbol at the first positions, different main symbols at the second positions e.g. \textbf{Sd-Bv} vs \textbf{Sw-Ah}: the dot product is $\frac{1}{10}$.
            \item[4.2.2)] Two common main symbols at the same positions e.g. \textbf{Ah-Bv} vs \textbf{Al-Bt}: the dot product is $\frac{1}{2}$ (consistent with property \textbf{2)} above, built in the algorithm itself).
            \item[4.2.3)] One common main symbol at crossed positions, the remaining positions have different main symbols e.g. \textbf{Ah-Bv} vs \textbf{Sw-Ap}: the dot product is $\frac{1}{5}$.
            \item[4.2.4)] Two common main symbols at crossed positions e.g. \textbf{Ah-Bv} vs \textbf{Bt-Al}: the dot product is $\frac{2}{5}$.
            \item[4.2.5)] One common main symbol at the second position, different main symbols at the first position e.g. \textbf{Ah-Bv} vs \textbf{Sd-Bt}: the dot product is $\frac{2}{5}$.
        \end{itemize}
    \end{itemize}
\end{itemize}

\subsection{Numerical Example}
Following is an example of how the graph embeddings are computed in a simplified setting with horizons from only four main classes: two classes with single representatives ($h_1 = \textbf{iC}$ and $h_2 = \textbf{Gor}$), two classes with multiple representatives ($h_3 = \textbf{Al}, h_4 = \textbf{Ael}, h_5 = \textbf{Acp}$ and $h_6 = \textbf{Bt}, h_7 = \textbf{Bs}, h_8 = \textbf{Bv}, h_9 = \textbf{Btv}$) that also have a mixture $h_{10} = \textbf{Al-Bv}$. This means that the embedding space dimension here is $d_E = 9$, with $\varphi(h_i) \in \mathbb{R}^9, \forall i = 1, ..., 10, N = 10$ (recall that mixtures do not add dimensionality to this space). $e_i \in \mathbb{R}^9$ denotes the unity vector with 1 at the $i$-th coordinate and 0's everywhere else.

\begin{enumerate}

    \item $ \varphi(h_1) = e_1 $ (\textbf{iC})

    \item $ \varphi(h_2) = e_2 $ (\textbf{Gor})

    \item $ \varphi(h_3) = e_3 $ (\textbf{Al})
    \begin{enumerate}[label*=\arabic*.]

        \item 
        $
        \left.
        \begin{array}{ll}
            \varphi(h_4)^T \cdot \varphi(h_1) \overset{!}{=} s(h_4, h_1) = 1 - 1 = 0 \\
            \varphi(h_4)^T \cdot \varphi(h_2) \overset{!}{=} s(h_4, h_2) = 1 - 1 = 0 \\
            \varphi(h_4)^T \cdot \varphi(h_3) \overset{!}{=} s(h_4, h_3) = 1 - \cfrac{1}{2} = \cfrac{1}{2}
        \end{array}
        \right \} \iff 
        \left\{
        \begin{array}{ll}
            \varphi(h_4)_1 \cdot 1 = 0 \\
            \varphi(h_4)_2 \cdot 1 = 0 \\
            \varphi(h_4)_3 \cdot 1 = \cfrac{1}{2}
        \end{array}
        \right \} \implies
        \implies \hat{\varphi}(h_4) = \bigg(0, 0, \cfrac{1}{2}\bigg)^T \\
        \implies \varphi(h_4)_4 = \sqrt{1 - ||\hat{\varphi}(h_4)||^2} = \sqrt{1 - \cfrac{1}{4}} = \cfrac{\sqrt{3}}{2} \\
        \implies \varphi(h_4) = \bigg(0, 0, \cfrac{1}{2}, \cfrac{\sqrt{3}}{2}, 0, ..., 0\bigg)^T
        $ (\textbf{Ael})
  
        \item 
        $
        \left.
        \begin{array}{ll}
            \varphi(h_5)^T \cdot \varphi(h_1) \overset{!}{=} s(h_5, h_1) = 1 - 1 = 0 \\
            \varphi(h_5)^T \cdot \varphi(h_2) \overset{!}{=} s(h_5, h_2) = 1 - 1 = 0 \\
            \varphi(h_5)^T \cdot \varphi(h_3) \overset{!}{=} s(h_5, h_3) = 1 - \cfrac{1}{2} = \cfrac{1}{2} \\
            \varphi(h_5)^T \cdot \varphi(h_4) \overset{!}{=} s(h_5, h_4) = 1 - \cfrac{1}{2} = \cfrac{1}{2}
        \end{array}
        \right \} \iff \\
        \iff \left\{
        \begin{array}{ll}
            \varphi(h_5)_1 \cdot 1 = 0 \\
            \varphi(h_5)_2 \cdot 1 = 0 \\
            \varphi(h_5)_3 \cdot 1 = \cfrac{1}{2} \\
            \varphi(h_5)_3 \cdot \varphi(h_4)_3 + \varphi(h_5)_4 \cdot \varphi(h_4)_4 = \cfrac{1}{2} \cdot \cfrac{1}{2} + \varphi(h_5)_4 \cdot \cfrac{\sqrt{3}}{2} = \cfrac{1}{2}
        \end{array}
        \right \} \implies
        \implies \hat{\varphi}(h_5) = \bigg(0, 0, \cfrac{1}{2}, \cfrac{\sqrt{3}}{6}\bigg)^T \\
        \implies \varphi(h_5)_5 = \sqrt{1 - ||\hat{\varphi}(h_5)||^2} = \sqrt{1 - (\cfrac{1}{4} + \cfrac{1}{12})} = \sqrt{1 - \cfrac{1}{3}} = \cfrac{\sqrt{6}}{3} \\
        \implies \varphi(h_5) = \bigg(0, 0, \cfrac{1}{2}, \cfrac{\sqrt{3}}{6}, \cfrac{\sqrt{6}}{3}, 0, ..., 0\bigg)^T
        $ (\textbf{Acp})
    \end{enumerate} 

    \item $ \varphi(h_6) = e_6 $ (\textbf{Bt})
    \begin{enumerate}[label*=\arabic*.]

        \item $ \varphi(h_7) = \bigg(0, ..., 0, \cfrac{1}{2}, \cfrac{\sqrt{3}}{2}, 0, 0\bigg)^T $ (\textbf{Bs}, analogous to $3.1$)

        \item $ \varphi(h_8) = \bigg(0, ..., 0, \cfrac{1}{2}, \cfrac{\sqrt{3}}{6}, \cfrac{\sqrt{6}}{3}, 0\bigg)^T $ (\textbf{Bv}, analogous to $3.2$)

        \item 
        Leaving out the dot products equaling 0 here:
        $
        \left.
        \begin{array}{ll}
            \varphi(h_9)^T \cdot \varphi(h_6) \overset{!}{=} s(h_9, h_6) = \cfrac{1}{2} \\
            \varphi(h_9)^T \cdot \varphi(h_7) \overset{!}{=} s(h_9, h_7) = \cfrac{1}{2} \\
            \varphi(h_9)^T \cdot \varphi(h_8) \overset{!}{=} s(h_9, h_8) = \cfrac{1}{2} \\
        \end{array}
        \right \} \iff \\
        \iff \left\{
        \begin{array}{ll}
            \varphi(h_9)_6 \cdot \varphi(h_6)_6 = \cfrac{1}{2} \\
            \varphi(h_9)_6 \cdot \varphi(h_7)_6 + \varphi(h_9)_7 \cdot \varphi(h_7)_7 = \cfrac{1}{2} \\
            \varphi(h_9)_6 \cdot \varphi(h_8)_6 + \varphi(h_9)_7 \cdot \varphi(h_8)_7 + \varphi(h_9)_8 \cdot \varphi(h_8)_8 = \cfrac{1}{2}
        \end{array}
        \right \} \implies \\
        \implies \left\{
        \begin{array}{ll}
            \varphi(h_9)_6 \cdot 1 = \cfrac{1}{2} \Rightarrow \varphi(h_9)_6 = \cfrac{1}{2} \\
            \cfrac{1}{2} \cdot \cfrac{1}{2} + \varphi(h_9)_7 \cdot \cfrac{\sqrt{3}}{2} = \cfrac{1}{2} \Rightarrow \varphi(h_9)_7 = \cfrac{\sqrt{3}}{6} \\
            \cfrac{1}{2} \cdot \cfrac{1}{2} + \cfrac{\sqrt{3}}{6} \cdot \cfrac{\sqrt{3}}{6} + \varphi(h_9)_8 \cdot \cfrac{\sqrt{6}}{3} = \cfrac{1}{2} \Rightarrow \varphi(h_9)_8 = \cfrac{\sqrt{6}}{12}
        \end{array}
        \right \} \implies \\
        \implies \hat{\varphi}(h_9) = \bigg(0, ..., 0, \cfrac{1}{2}, \cfrac{\sqrt{3}}{6}, \cfrac{\sqrt{6}}{12}\bigg)^T \in \mathbb{R}^8 \\
        \implies \varphi(h_9)_9 = \sqrt{1 - ||\hat{\varphi}(h_9)||^2} = \sqrt{1 - (\cfrac{1}{4} + \cfrac{1}{12} + \cfrac{1}{24})} = \sqrt{1 - \cfrac{3}{8}} = \cfrac{\sqrt{10}}{4} \\
        \implies \varphi(h_5) = \bigg(0, ..., 0, \cfrac{1}{2}, \cfrac{\sqrt{3}}{6}, \cfrac{\sqrt{6}}{12}, \cfrac{\sqrt{10}}{4}\bigg)^T
        $ (\textbf{Btv})
    \end{enumerate}

    \item $ \varphi(h_{10}) = \cfrac{1}{3} \varphi(h_3) + \cfrac{2}{3} \varphi(h_8) $ (\textbf{Al-Bv})
    
    Set $ \varphi(h_{10}) \leftarrow \varphi(h_{10}) / || \varphi(h_{10}) ||$
\end{enumerate}

\FloatBarrier
\section{Compute Resources and Hyperparameters}
\label{sec:appendix_exp_details}

In this section, we provide additional details on the computational resources and hyperparameters used in our experiments.

\paragraph{Experiment Compute Settings.}
The experiments were conducted on our internal compute cluster, shared with other researchers. We used NVIDIA B200 GPUs with a total of 180GB of VRAM each. Each experiment was run on a single GPU, without any distributed training or data parallelism. The training was done using the PyTorch framework (PyTorch 2.7.0 with CUDA 12.8) and the experiment results were tracked on Weights \& Biases (\url{https://wandb.ai/site}). For the end-to-end SoilNet models, the training took on average 45 minutes per experiment and an average memory usage of 150GB of VRAM on the GPU. We report 8 experiments for the end-to-end SoilNet models, which makes a total of 6 hours of training time. For the individual tasks the memory usage as well as the training time was lower, with an average of 25 minutes per experiment and an average memory usage of 100GB of VRAM on the GPU. We report 8 experiments for the individual tasks, which makes a total of 2.5 hours of training time. With the unreported experiments neglected, the total training time for all experiments is 8.5 hours.

\paragraph{Hyperparameters.}
In our experiments, we used common default hyperparameters with minimal adjustments from preliminary trials, without performing extensive searches. \autoref{tab:hyperparameters_general} shows the hyperparameters used for all experiments in this paper. For the additional hyperparameters specific to each experiment and model configuration, the readers may consult our code repository, where we list them at the end of the experiment files.

\begin{table*}[pos=h]
    \small
    \centering
    \caption{Hyperparameters used for all experiments.}
    \label{tab:hyperparameters_general}
    \begin{tabular}{lc}
        \toprule
        Hyperparameter & Value \\
        \midrule
        Batch size & 8 \\
        Number of workers & 16 \\
        Optimizer & AdamW \\
        Learning rate & 0.0001 \\
        Dropout & 0.1 \\
        Weight decay & 0.01 \\
        Maximum epochs & 100 \\
        Early stopping patience & 5 \\
        \bottomrule
    \end{tabular}
\end{table*}

\section{Full Metrics Tables for Individual Tabular Prediction}
\label{sec:appendix_simple_tasks}

In \autoref{subsubsec:task2}, \autoref{tab:aggregated_metrics_summary} shows the aggregated metrics for the categorical features for the two configurations of the individual tabular predictor (Task 2). Here in \autoref{tab:simple_tabular_full}, we present the full metrics tables for predicting the morphological horizon-specific features. Both configurations use LSTM as tabular prediction modules for the six tabular horizon features. They differ in the segment encoder: one uses the custom PatchCNN and the other the ResNet-based segment encoder (see \autoref{subsec:soilnet}). The ResNet-based tabular predictor outperforms the PatchCNN counterpart.

\begin{table*}[pos=h]
\small 
  \caption{\textbf{Full tables for individual tabular prediction metrics (Task 2).} Metrics on the test set. 
  \\ TP = Tabular Predictor, \{PatchCNN, ResNet\} = segment encoder, LSTM = tabular prediction module, Acc. = Accuracy, Prec. = Precision, Rec. = Recall.}
  \label{tab:simple_tabular_full}
  \centering
  \resizebox{\textwidth}{!}{
  \begin{tabular}{lcccccccc}
    \toprule
     & \multicolumn{8}{c}{TP\_PatchCNN\_LSTM} \\ 
    \cmidrule(r){2-9}
    Feature & MSE & Acc. (\%) & F1 (\%) & Prec. (\%) & Rec. (\%) & Acc.@3 (\%) & Prec.@3 (\%) & Rec.@3 (\%) \\
    \midrule
    Stones       & 6.11 & -    & -    & -    & -    & -    & -    & -    \\
    Soil Texture & -    & 34.55 & 17.02 & 25.72 & 17.25 & 65.56 & 52.08 & 36.96 \\
    Soil Color   & -    & 24.15 & 2.38  & 4.55  & 3.49  & 49.98 & 19.48 & 10.11 \\
    Carbonate    & -    & 72.57 & \textbf{21.03} & \textbf{39.79} & \textbf{22.88} & 92.06 & 74.37 & \textbf{52.84} \\
    Humus        & -    & 54.57 & 32.97 & 34.75 & 33.00 & 91.20 & 87.64 & 67.14 \\
    Rooting      & -    & 36.79 & 27.70 & 32.56 & 30.33 & 74.54 & 74.81 & 70.55 \\
    \bottomrule
    \toprule
     & \multicolumn{8}{c}{TP\_ResNet\_LSTM} \\ 
    \cmidrule(r){2-9}
    Feature & MSE & Acc. (\%) & F1 (\%) & Prec. (\%) & Rec. (\%) & Acc.@3 (\%) & Prec.@3 (\%) & Rec.@3 (\%) \\
    \midrule
    Stones       & \textbf{1.30} & -        & -        & -        & -        & -        & -        & -        \\
    Soil Texture & -             & \textbf{34.99} & \textbf{20.20} & \textbf{25.81} & \textbf{21.23} & \textbf{69.35} & \textbf{67.62} & \textbf{48.77} \\
    Soil Color   & -             & \textbf{29.37} & \textbf{5.53}  & \textbf{6.77}  & \textbf{5.75}  & \textbf{57.74} & \textbf{20.08} & \textbf{14.18} \\
    Carbonate    & -             & \textbf{78.38} & 20.29 & 31.06 & 21.43 & \textbf{92.73} & \textbf{81.68} & 49.58 \\
    Humus        & -             & \textbf{56.16} & \textbf{40.82} & \textbf{42.39} & \textbf{45.26} & \textbf{92.66} & \textbf{88.82} & \textbf{84.61} \\
    Rooting      & -             & \textbf{39.04} & \textbf{33.93} & \textbf{41.19} & \textbf{33.92} & \textbf{77.14} & \textbf{81.87} & \textbf{73.55} \\
    \bottomrule
  \end{tabular}
  }
\end{table*}

\section{Confusion Matrices for Individual Horizon Classification}
\label{sec:appendix_confusion_matrices}

\begin{figure*}[!htbp]
    \centering
    \subfloat[HP\_ResNet\_LSTM\_CE\label{subfig:cm_ce}]{\includegraphics[width=0.45\textwidth]{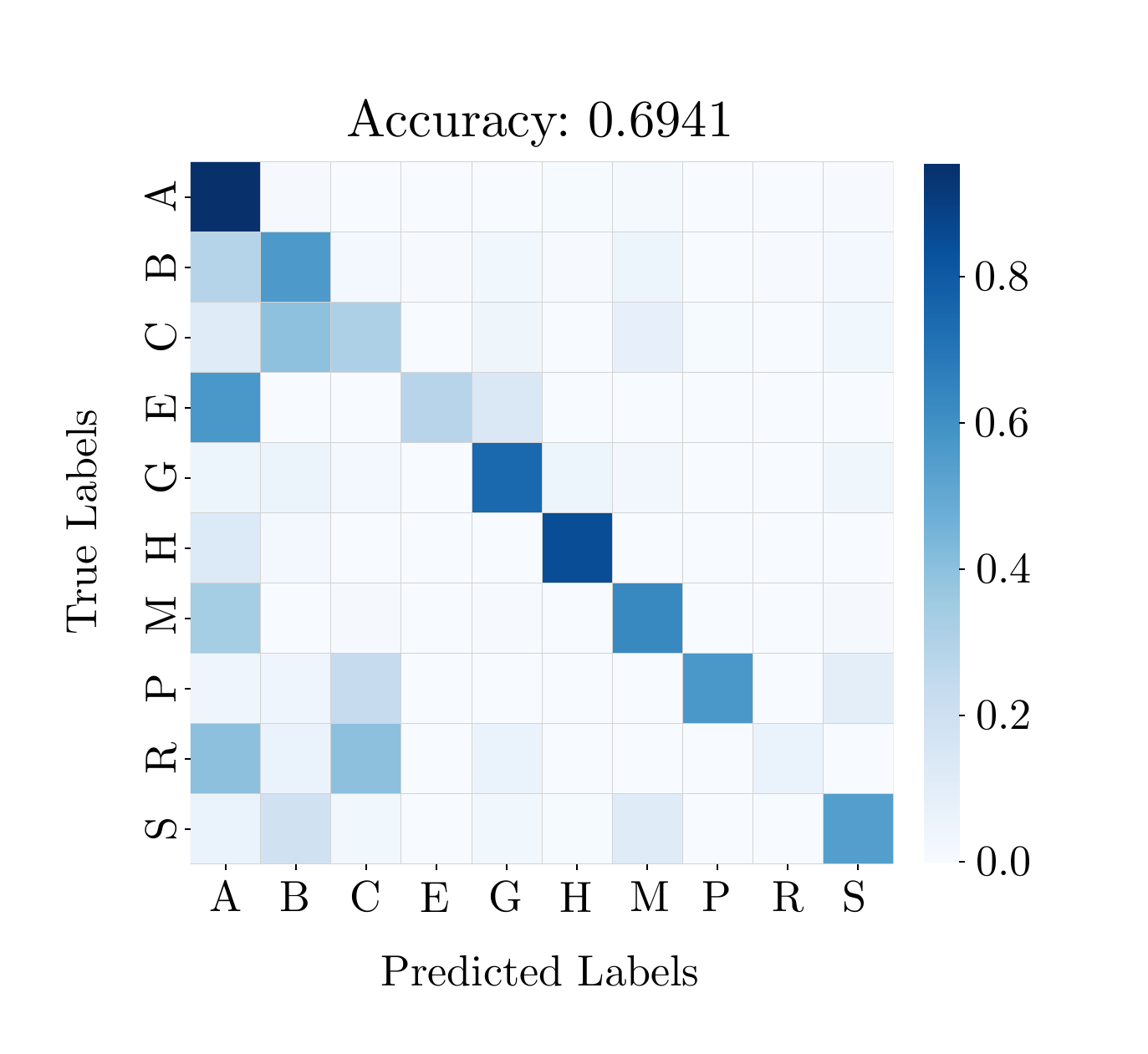}}
    \hspace{0.05\textwidth} 
    \subfloat[HP\_ResNet\_LSTM\_Emb\label{subfig:cm_emb}]{\includegraphics[width=0.45\textwidth]{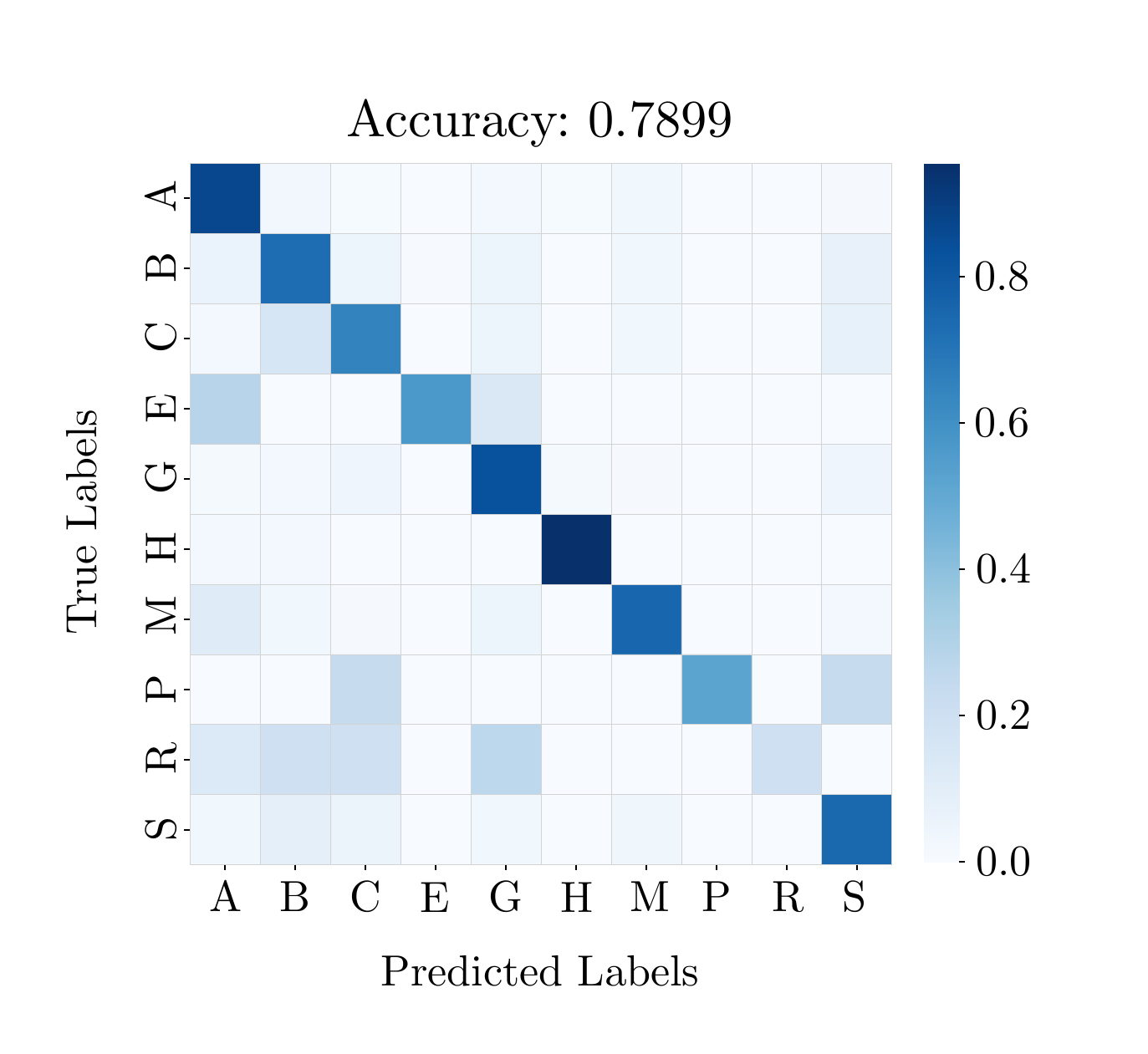}}
    \caption{\textbf{An embedding-based horizon classifier predicts the main symbol more accurately than a classifier trained on one-hot encodings.} The confusion matrices are aggregated on the main symbol hierarchy.\ 
    HP = Horizon Predictor, ResNet = segment encoder (see \autoref{subsec:soilnet}), LSTM = horizon classification module, Emb = embedding loss, CE = cross entropy.}
    \label{fig:confusion_matrices}
\end{figure*}

We show here only the confusion matrices aggregated on the main symbol level for two individual Task 3 solvers (\autoref{fig:confusion_matrices}). The full confusion matrices for these two examples with all 99 horizon labels can be found in our repository, as well as matrices for other horizon predictors.

\FloatBarrier 

\end{document}